\documentclass{article}

\usepackage{microtype}
\usepackage{graphicx}
\usepackage{booktabs} % for professional tables
\usepackage[table]{xcolor} % 导入xcolor包，并启用table选项
\usepackage{colortbl} % 导入colortbl包
\usepackage{dashrule}
\usepackage{hyperref}

\usepackage[accepted]{icml2025}

\usepackage{amsmath}
\usepackage{amssymb}
\usepackage{mathtools}
\usepackage{amsthm}
\usepackage{multirow}
\usepackage{array}
\usepackage{diagbox}
\usepackage{subcaption}  % 用于子表格

\usepackage[capitalize,noabbrev]{cleveref}

\theoremstyle{plain}

\theoremstyle{definition}

\theoremstyle{remark}

\usepackage[textsize=tiny]{todonotes}
\definecolor{mygray}{gray}{.92}

\icmltitlerunning{A Token-level Text Image Foundation Model for Document Understanding}

\begin{document}

\twocolumn[
\icmltitle{A Token-level Text Image Foundation Model for Document Understanding}



\icmlsetsymbol{equal}{*}

\begin{icmlauthorlist}
\icmlauthor{Tongkun Guan}{yyy,equal}
\icmlauthor{Zining Wang}{comp,equal}
\icmlauthor{Pei Fu}{comp}
\icmlauthor{Zhengtao Guo}{sch}
\icmlauthor{Wei Shen}{yyy}
\icmlauthor{Kai Zhou}{comp}
\icmlauthor{Tiezhu Yue}{comp}
\icmlauthor{Chen Duan}{comp}
\icmlauthor{Hao Sun}{ggg}
\icmlauthor{Qianyi Jiang}{comp}
\icmlauthor{Junfeng Luo}{comp}
\icmlauthor{Xiaokang Yang}{yyy}
% \icmlauthor{}{sch}
% \icmlauthor{}{sch}
\url{https://github.com/Token-family/TokenFD}
\end{icmlauthorlist}

\icmlaffiliation{yyy}{MoE Key Lab of Artificial Intelligence, AI Institute, Shanghai Jiao Tong University, Shanghai, China}
\icmlaffiliation{comp}{Meituan}
\icmlaffiliation{sch}{Beijing Institute of Technology, Beijing, China}
\icmlaffiliation{ggg}{Institute of Automation, Chinese Academy of Sciences, Beijing, China}

% \icmlcorrespondingauthor{Tongkun Guan}{gtk0615@sjtu.edu.cn}
% \icmlcorrespondingauthor{Wei Shen}{wei.shen@sjtu.edu.cn}

\icmlkeywords{Machine Learning, ICML}

\vskip 0.3in
]




\newcolumntype{L}[1]{>{\raggedright\arraybackslash}p{#1}}
\newcolumntype{C}[1]{>{\centering\arraybackslash}p{#1}}
\newcolumntype{R}[1]{>{\raggedleft\arraybackslash}p{#1}}

\begin{abstract}
In recent years, general visual foundation models (VFMs) have witnessed increasing adoption, particularly as image encoders for popular multi-modal large language models (MLLMs). However, without semantically fine-grained supervision, these models still encounter fundamental prediction errors in the context of downstream text-image-related tasks, \emph{i.e.}, perception, understanding and reasoning with images containing small and dense texts. To bridge this gap, we develop \textbf{TokenFD}, the first token-level visual foundation model specifically tailored for text-image-related tasks, designed to support a variety of traditional downstream applications. To facilitate the pretraining of TokenFD, we also devise a high-quality data production pipeline that constructs the first token-level image text dataset, \textbf{TokenIT}, comprising 20 million images and 1.8 billion token-mask pairs. Furthermore, leveraging this foundation with exceptional image-as-text capability, we seamlessly replace previous VFMs with TokenFD to construct a document-level MLLM, \textbf{TokenVL}, for VQA-based document understanding tasks.
Finally, extensive experiments demonstrate the effectiveness of TokenFD and TokenVL.
\end{abstract}
\section{Introduction}
Text image acts as a crucial medium for information transmission in everyday life. The precise interpretation of these images significantly enhances the automation of information processes, including text recognition, retrieval, segmentation, and understanding.

\begin{figure}[t]
    \centering
    \begin{minipage}{0.5\textwidth}
        \centering
        \includegraphics[width=0.9\textwidth]{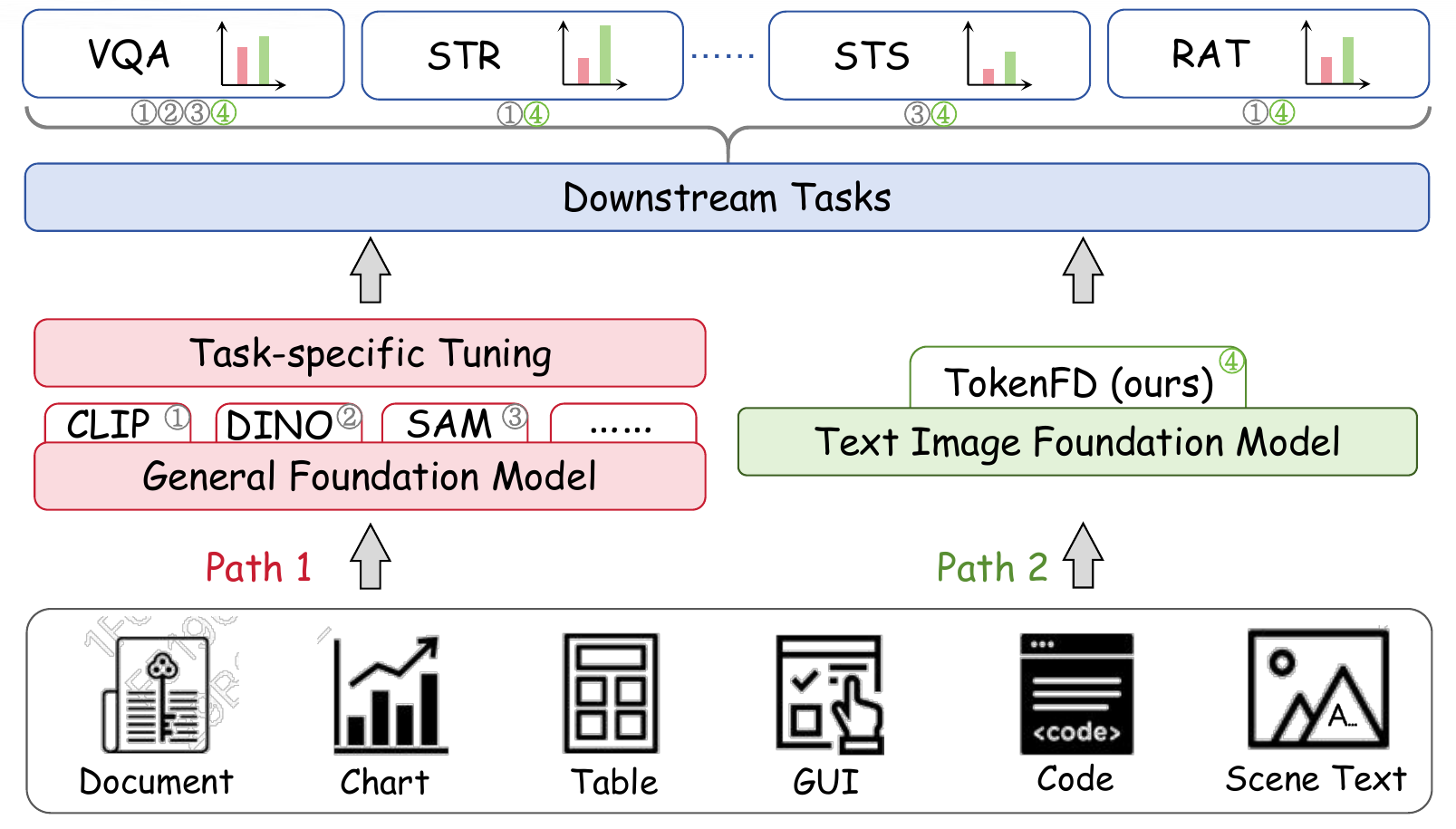} % 替换为你的图像文件路径
        % \caption{Example Image}
        % \label{fig:example}
    \end{minipage}
    % \vspace{1em} % 添加一些垂直间距
    \begin{minipage}{0.5\textwidth}
        \centering
        \scalebox{0.8}{
    \begin{tabular}{l|L{17mm}|L{12mm}C{7mm}C{7mm}}
    \toprule
     VFM & Granularity &Dataset & \#Image &\#Pairs \\
     \midrule
     CLIP~\cite{CLIP} & image-level &WIT400M  &400M &0.4B \\
     DINO~\cite{DINO} & image-level &ImageNet &14M &- \\
     SAM~\cite{SAM} & pixel-level &SA1B &11M &1.1B \\
     \midrule
     TokenFD & token-level &TokenIT &20M &1.8B \\
     \bottomrule
\end{tabular}
    }
    % \vspace{-0.5em}
    \caption{For different tasks, previous works select different VFMs from general foundation models (path 1). In contrast, we develop a unified token-level foundation model, \textbf{TokenFD}, specifically tailored for text-image-related tasks (path 2).
    TokenFD is trained on a substantial self-built dataset, \textbf{TokenIT}, comprising 20 million images and 1.8 billion token-mask pairs. This well-learned model is capable of supplanting other VFMs in related downstream tasks.
    }
    \label{thumbnail}
    \end{minipage}
    % \vspace{-1.5em}
\end{figure} 

With the trend towards these tasks unification and the advancement of multi-modal large language models (MLLMs), visual foundation models (VFMs) have garnered considerable attention due to their broad capabilities in providing visual understanding for these downstream vision tasks~\cite{covert2024locality}. For instance, popular general models  CLIP~\cite{CLIP}, DINO~\cite{DINO}, and SAM~\cite{SAM} are widely adapted for text-image-related tasks to achieve performance gains through LoRA/adapter tuning~\cite{HISAM}, prompt learning~\cite{TCM}, and learnable position interpolation technology. Additionally, CLIP and SigLIP~\cite{SigLIP} have also proven effective as visual encoders for MLLMs in concurrent studies~\cite{MM1,cambrian}.

However, these VFMs, trained with image-level supervision, are not optimal for processing fine-grained dense prediction tasks~\cite{yu2024texthawk2}, such as document understanding with densely packed and small visual texts. Although several works attempt to incorporate SAM as an additional high-resolution encoder~\cite{vary,mousi} or combine other expert models~\cite{sphinx}, these dual or more complex VFM combinations result in a doubling of the number of tokens, which is costly and lack flexibility. Furthermore, to the best of our knowledge, there is currently almost no fine-grained text image foundation model with token granularity,  specifically tailored for extracting robust and general visual text semantic feature representations.

In this work, we close the gap and explore the potential of the text image foundation model at a large scale.
Leveraging the vast amounts of publicly available data, we develop a high-quality data production pipeline that constructs the first token-level image text dataset, named \textbf{TokenIT}, comprising 20 million images and 1.8 billion token-mask pairs. Specifically, we begin by extracting text transcriptions and text masks for each sample. Subsequently, we split each text transcription into several tokens (BPE-level subwords) using a tokenizer~\cite{internvl} and obtain their corresponding BPE token masks. The number of token-mask pairs ultimately constructed is 4.5 times that of CLIP and 0.7B more than SAM as summarized in Figure \ref{thumbnail}.

Leveraging the self-constructed TokenIT dataset, we further propose the first token-level text image foundation model, named \textbf{TokenFD}, designed to support a wide array of text-image-related downstream tasks. To achieve \emph{image-as-text} semantic alignment, token-level visual embeddings are aligned with token-level language embeddings for positive token-mask pairs, meanwhile ensuring that negative pairs remain distinct within the embedding space. Specifically, each token-level visual embedding is derived through a mean pooling operation applied to the visual image features within a corresponding token mask; each token-level language embedding is produced via a straightforward token embedding layer, obviating the need for a complex text encoder like CLIP. 

The \emph{image-as-text} semantic attributes, aligned at the VFM level, effectively bridge the gaps between visual and language modalities. This approach creates a unified sequence representation that can be seamlessly integrated into any large language model (LLM) for popular MLLM tasks. Building upon this foundation, we propose a document-level MLLM, named \textbf{TokenVL}, which further enhances spatially visual-language token alignment at the LLM level for document understanding in Visual Question Answering (VQA) tasks. Additionally, we freeze the weights of the TokenFD model to facilitate other downstream applications, including text segmentation, text retrieval, and end-to-end text recognition tasks.

Overall, the main contributions are summarized as follows:
\begin{itemize}
    \item[1)] The first token-level image text dataset (TokenIT) is proposed, which consists of 20M images and 1.8B high-quality token-mask pairs.
    \item[2)] The first token-level text image foundation model, TokenFD, is proposed to support various downstream tasks, including text segmentation, text retrial, and text understanding.
    \item[3)] The \emph{image-as-text} semantic capability inspires us to develop TokenVL, a VQA-based MLLM tailored for document perception, understanding, and reasoning.
    \item[4)] Extensive experiments demonstrate the effectiveness of our proposed TokenFD and TokenVL. Specifically, TokenFD shows exceptional "zero-shot" capabilities and flexibility compared to other VFMs, such as CLIP, SAM, and InternViT2.5. TokenVL with 8B parameters, incorporating TokenFD as the VFM, achieves performance gains of 38 on the OCRBench task and an average of 8.8\% across ten document VQA tasks. Similarly, TokenVL with 2B parameters results in performance gains of 17 on the OCRBench task and an average of 13.34\% on the ten VQA tasks.
\end{itemize}
\section{Related Work}
\label{sec:formatting}

% \subsection{Visual Foundation Models}
\noindent \textbf{Visual Foundation Models.}
Visual foundation models (VFMs) are a vitally important component, which serves various downstream tasks, such as semantic segmentation~\cite{shen2023survey}, optical character recognition~\cite{guan2022industrial,guan2025bridging,guan2025posformer}, object detection~\cite{liu2025grounding}, and remote sensing~\cite{hong2024spectralgpt}.  
Notably, Radford et al.~\cite{CLIP} introduce CLIP to align visual and language modalities through contrastive learning from large-scale image-text pairs. SigLIP~\cite{SigLIP} demonstrate that a simple sigmoid loss can be more effective than a contrastive loss. Caron et al.~\cite{DINO} propose DINO, a method for self-supervised learning of image features without labeled data, utilizing self-distillation. However, several studies have observed that these image-level supervised paradigms often encounter basic perceptual errors and fail to capture localized features necessary for dense prediction tasks.
Kirillov et al.~\cite{SAM} introduce the pixel-level SAM, ushering in a new era of segmenting virtually anything. 
Despite the model's prominence in segmentation tasks, its limited semantic capabilities constrain its applicability to tasks requiring deeper understanding and reasoning. Recently, with the advancement of multimodal large language models (MLLMs) and the trend towards task unification, building more suitable VFMs has become increasingly important. 

\begin{figure*}[t]
    \centering
    \includegraphics[width=0.97\linewidth]{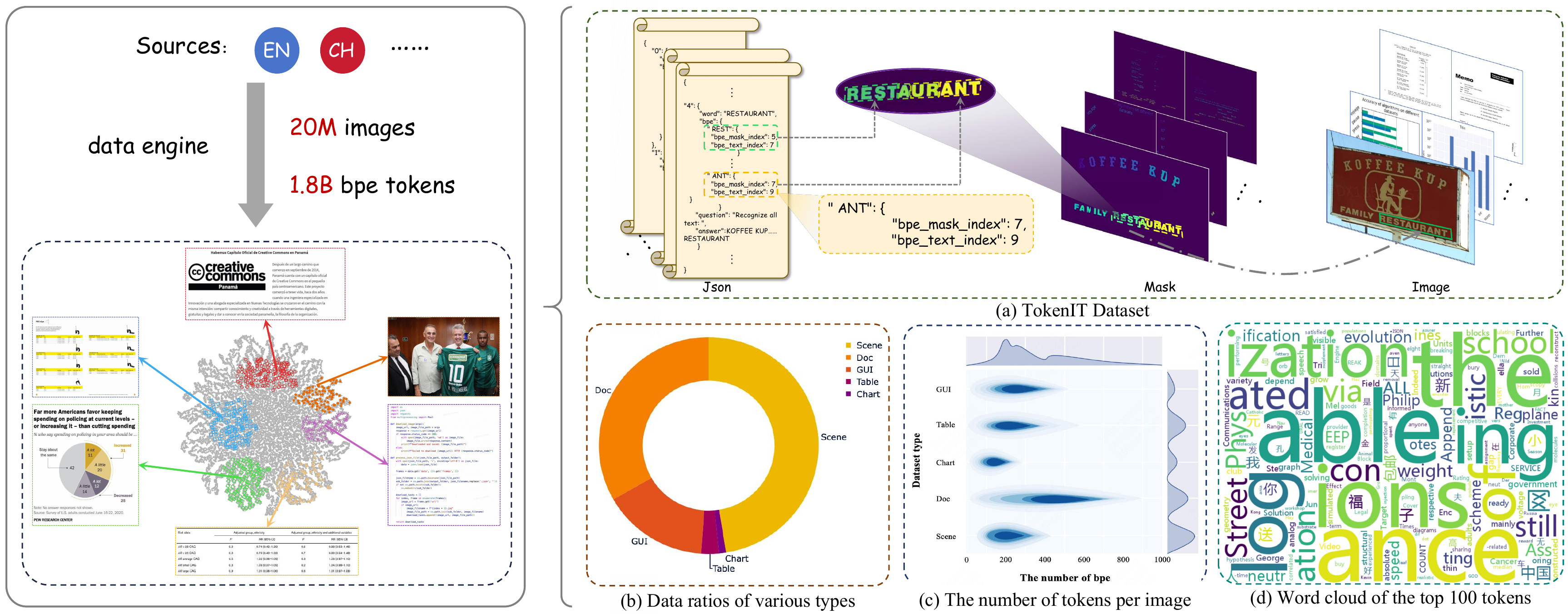}
    \caption{
    An overview of the self-constructed token-level TokenIT dataset,  comprising 20 million images and 1.8 billion text-mask pairs. (a) provides a detailed description of each sample, including the raw image, a mask, and a JSON file that records BPE token information. 
    We also count (b) the data distribution, (c) the number of selected BPE tokens, and (d) a word cloud map highlighting the top 100 BPE tokens.
    }
    \label{fig:dataset}
    % \vspace{-0.5em}
\end{figure*}

\noindent \textbf{MLLMs for Document Understanding.}
Multimodal Large Language Models (MLLMs) connect the powerful Visual Foundation Model and Large Language Model to facilitate perception, understanding, and reasoning, which generate coherent texts through visual question answering. Recent advancements have empowered MLLMs to extract meaningful information from text images for Visual Document Understanding (VDU) tasks. Specifically, these methods can be roughly categorized into two types: OCR-dependent MLLMs~\cite{LayTextLLM,MOAI,Cream,Instructdoc,Doclayllm} and OCR-free MLLMs~\cite{internvl,minigpt,MiNi-Monkey,Monkey,Mplug-Docowl2,UReader}.
OCR-dependent MLLMs utilize an external OCR engine to extract text information and merge the generated results into MLLMs, which brings excessive auxiliary tokens.
In contrast, OCR-free MLLMs have sought to simplify this process by predicting question-driven outputs directly. They incorporate task-specific modules for enhancing the capabilities of Document MLLMs, including high-resolution image processing~\cite{Monkey,UReader,mplug-docowl1.5,docpedia}, efficient token compression~\cite{zhang2024token,Mplug-Docowl2,yu2024texthawk2}, and refined attention mechanisms~\cite{MiNi-Monkey,VisualCoT}.
Despite these achievements, existing OCR-free models still struggle to capture fine-grained textual content within images. We speculate that this limitation is caused by the VFMs utilized in large multimodal models. Therefore, we propose the first token-level text image foundation model for visual document understanding tasks. This model aims to bridge the visual-language modality gap by ensuring that the semantic descriptions of each BPE token of visual texts in an image correspond accurately to those of language texts.
\begin{figure*}[t]
    \centering
    \includegraphics[width=0.95\linewidth]{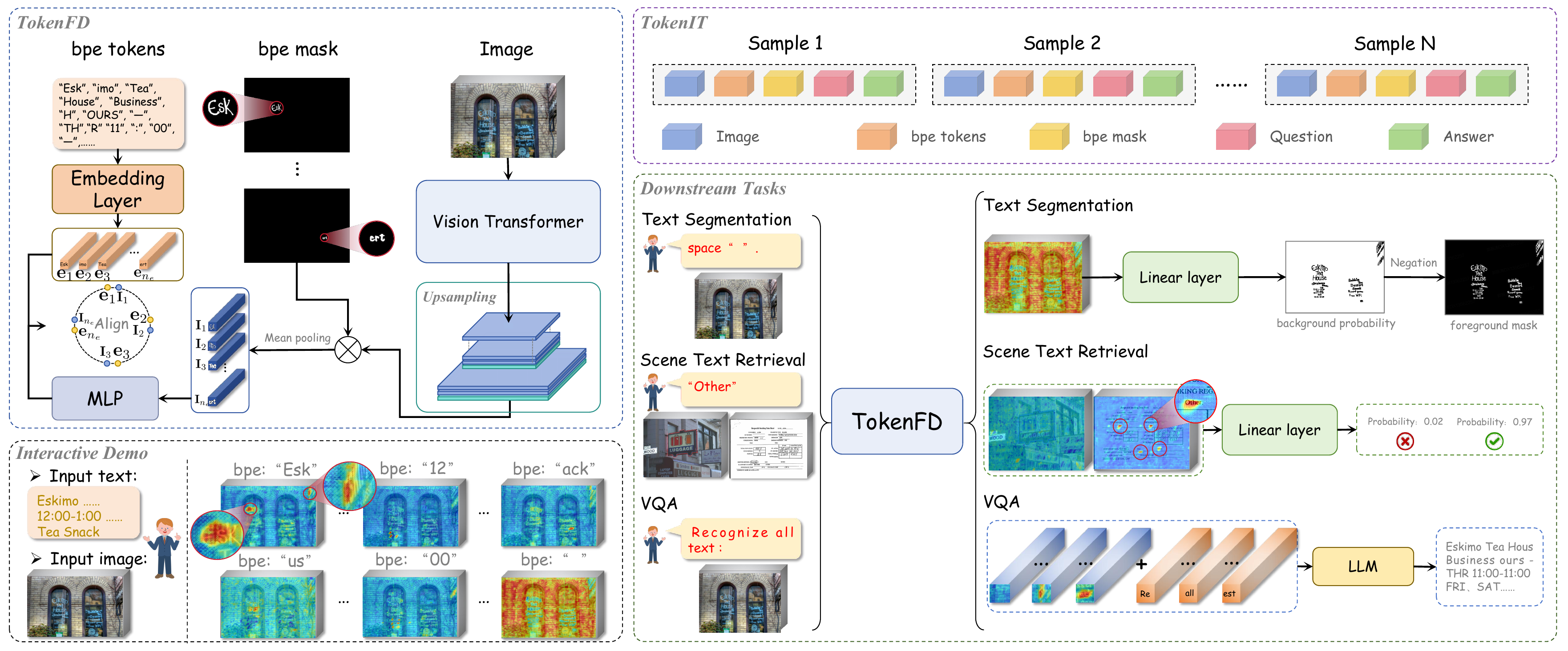}
    % \vspace{-1.0em}
    \caption{An overview of the proposed TokenFD, where the token-level image features and token-level language features are aligned within the same semantic space. This ``image-as-text'' alignment seamlessly facilitates user-interactive applications, including text segmentation, retrieval, and visual question answering.}
    \label{fig:arch}
    % \vspace{-1.em}
\end{figure*}

\section{TokenIT Dataset}
In the OCR community, there are almost no datasets of image-text pairs with token granularity, where each language token (split by the BPE tokenizer) aligns precisely with its corresponding image location. However, this type of dataset could effectively enhance the fine-grained perception of VFMs and assist MLLMs in bridging the modality gap between visual and language embeddings. To fill this gap, we curate a Token-level Image Text dataset, TokenIT. 

Specifically, to construct a robust and comprehensive TokenIT dataset, we collect various types of data, including natural scene text images, documents (PDF, receipt, letter, note, report, \emph{etc.}), tables, charts, code, and GUI. Finally, three rounds of inspections are conducted to minimize labeling errors, a process that took four months to develop the first token-level image text dataset (TokenIT), which includes 20 million images and 1.8 billion token-mask pairs.

As depicted in Figure \ref{fig:dataset} (a), each sample in this dataset includes a raw image, a mask image, and a JSON file. The JSON file provides the question-answer pairs and several BPE tokens randomly selected from the answer, along with the ordinal number of each BPE token in the answer and its corresponding pixel value on the mask image. Consequently, each BPE token corresponds one-to-one with a pixel-level mask.
The data ratios are summarized in Figure \ref{fig:dataset} (b). Figure \ref{fig:dataset} (c) and (d) further provide the number distribution of tokens per image type and a word cloud of the top 100 tokens, respectively. More specific details are introduced in Supplementary Material.
\section{Methodology}

% \subsection{Overall}
\noindent \textbf{Overall.}
To better describe our method, we define each sample $\mathcal{S}$ of our TokenIT dataset:
{\setlength\abovedisplayskip{0.5pt}
\setlength\belowdisplayskip{0.pt}
\begin{equation}
  \begin{cases}
    \begin{aligned}
        \mathcal{S} &= \{\mathbf{X}, \mathbf{M}, \mathcal{E}, \mathcal{Q}, \mathcal{A}\},\\
        \mathbf{M} &\Rrightarrow \{\textbf{M}_{1},...,\textbf{M}_{n_{e}}\},\\
        \mathcal{E} &= \{e_{1},...,e_{n_{e}} \},\\
        \mathcal{Q} &= \{q_{1},...,q_{n_{q}} \},\\
        \mathcal{A} &= \{a_{1},...,a_{n_{a}} \}, \\
    \end{aligned}
    \end{cases}
    \label{eq:sample}
\end{equation}
}

\noindent where $\mathbf{X}$ is an input image. $\mathcal{Q}$ and $\mathcal{A}$ denote the tokenized question and answer, respectively, processed using a BPE tokenizer~\cite{internvl}.
$\mathbf{M}$ refers to the mask image, which is divided into $n_{e}$ BPE token masks $\{\textbf{M}_{1},...,\textbf{M}_{n_{e}}\}$, according to the pixel value (recorded in the JSON file) of each BPE token on the mask image.
% These BPE token masks are created based on the pixel values corresponding to each BPE token on the mask image, as recorded in a JSON file. 
Consequently, for any BPE token ($e_{i}$ in $\mathcal{E}$), the pixel value at its specific position in the mask image 
$\textbf{M}_{i}$ is set to 1, with all other positions set to 0. Notably, $\mathcal{E}$ is a subset consisting of $n_{e}$ BPE tokens, which are randomly selected from $\mathcal{A}$.

Utilizing the TokenIT dataset with 1.8B token-mask pairs, we construct the first token-level OCR foundation model (TokenFD) by token-level \emph{image-as-text} alignment. For VQA-based document understanding downstream tasks, we employ the well-learned foundation model to construct an MLLM (TokenVL), which includes the following stages: 1) LLM-guided Token Alignment; 2) Supervised Instruction Tuning. Besides, we also freeze the foundation model (unless otherwise stated) to conduct other text-related downstream tasks, including text segmentation, text retrieval, and text understanding.

\subsection{TokenFD}
Although existing VFMs produce good representations for zero-shot or fine-tuning tasks, they still encounter significant challenges in processing fine-grained tasks, such as document scenarios with densely packed small texts. Thus a suitable VFM that is tailored for text images is in demand. In light of this, we construct the first token-level VFM, which fills the gap in the field. Concretely, the pre-training process is formulated as follows:

The input image $\mathbf{X}\in \mathbb{R}^{H\times W\times 3}$ is first fed into a ViT-based visual encoder $f(\cdot)$ to extract image features $\mathbf{F}\in \mathbb{R}^{\frac{H}{p}\times \frac{W}{p}\times C}$, where $p$ is the patch size, set to 14 by default. A simple two-layer deconvolution is then applied to the image feature $\mathbf{F}$ to enlarge the feature resolution. Subsequently, a linear layer ($\mathbb{R}^{C} \xrightarrow{} \mathbb{R}^{D}$) is applied to expand to the same embedding dimension as the language embedding layer. The processed image feature is denoted as $\mathbf{\tilde{F}}\in \mathbb{R}^{\frac{4\times H}{p}\times \frac{4\times W}{p}\times D}$. 

Next, given all BPE token-mask pairs $\mathcal{B} = \{(e_{1}, \mathbf{M}_{1}), (e_{2}, \mathbf{M}_{2}), ..., (e_{n_{e}}, \mathbf{M}_{n_{e}}) \}$ corresponding to the input image, the pre-training objective encourages embeddings of matching pairs $\{(\mathbf{e}_{1}, \mathbf{t}_{1}), (\mathbf{e}_{2}, \mathbf{t}_{2}), ..., (\mathbf{e}_{n_{e}}, \mathbf{t}_{n_{e}})\}$ to align with each other, where $\mathbf{e}_{i}\in \mathbb{R}^{D}$ is the token embeddings of $e_{i}$. The associate token-level visual features $\mathbf{t}_{i}\in \mathbb{R}^{D}$ are yielded by a mean-pooling operation:
{\setlength\abovedisplayskip{1pt}
\setlength\belowdisplayskip{1pt}
\begin{equation}
    \begin{aligned}
    \mathbf{t}_{i} = \frac{1}{\sum_{x,y}\mathrm{\texttt{BI}(M_{i})}^{(x,y)}}\sum_{x,y}\mathrm{\texttt{BI}(M_{i})}^{(x,y)}\mathbf{\tilde{F}}^{(x,y)},\\
    \end{aligned}
\end{equation}
}
where $\texttt{BI}(\cdot)$ refers to the bilinear interpolation operation to match the feature resolution of $\mathbf{\tilde{F}}$. The coordinate $(x, y)$ indicates a point on the $x$-axis and $y$-axis, respectively. 
Finally, without requiring a complex text encoder like CLIP-Text, we adopt a simple and learnable token embedding layer to align the visual-language modality at the token level. Specifically, following the previous works~\cite{SigLIP, zhang2023faster, guan2023self}, the objectives are to minimize:
{\setlength\abovedisplayskip{0pt}
\setlength\belowdisplayskip{0pt}
\begin{equation}
  \begin{cases}
    \begin{aligned}
        \mathcal{L}_{dis} &= \frac{1}{|\mathcal{B}|}\frac{1}{D}\sum_{i=1}^{|\mathcal{B}|}\sum_{j=1}^{D} \lvert e_{i}^{j} - \mathrm{t}_{i}^{j} \rvert,\\
        \mathcal{L}_{sim} &= \frac{1}{|\mathcal{B}|}\sum_{i=1}^{|\mathcal{B}|}\big( 1 - \frac{\mathbf{e}_{i} \cdot \mathbf{t}_{i}}{\|\mathbf{e}_{i}\| \|\mathbf{t}_{i}\|} \big),\\
        \mathcal{L}_{sig} &= - \frac{1}{|\mathcal{B}|} \sum_{i=1}^{|\mathcal{B}|} \sum_{j=1}^{|\mathcal{B}|} \underbrace{\log \frac{1}{1 + e^{z_{ij} (-k \mathbf{e}_i \cdot \mathbf{t}_j + b)}}}_{\mathcal{L}_{sig}^{ij}}, \\
        % \mathcal{L}_{all} &= \mathcal{L}_{dis} + \mathcal{L}_{sim} + \mathcal{L}_{sig}, \\
    \end{aligned}
    \end{cases}
    \label{eq:alignment}
\end{equation}
}

\noindent where $k$ and $b$ are learnable parameters, and we initialize $k$ and $b$ to \texttt{log\,10} and \texttt{-10}, respectively. 
The label $z_{ij}$ indicates whether the token-level visual feature $\mathbf{t}_{i}$ and token embedding $\mathbf{e}_{j}$ are a pair, being \texttt{1} if they are paired and \texttt{-1} otherwise.

After pre-training, the input image's visual embeddings and corresponding text embeddings share the same feature space, achieving \emph{image-as-text} semantic alignment. This alignment facilitates seamless image-text interaction, \emph{i.e.,} inputting text to highlight the corresponding area in the image (as illustrated in the ``Interactive Demo'' area of Figure \ref{fig:arch}), along with other derivative downstream tasks. More visualization examples are presented in Supplementary Materials.

\subsection{TokenVL}
The \emph{image-as-text} semantic attributes inherently bridge the gaps between visual and language modalities, creating a unified sequence representation that LLM can effectively understand. Inspired by this, we employ the TokenFD as the visual foundation model and further develop an MLLM, named TokenVL, tailored for document understanding. Following the previous training paradigm~\cite{internvl,lv2023kosmos,vary,mplug-docowl1.5}, TokenVL also includes two stages: 1) LLM-guided Token Alignment Training for text parsing tasks and 2) Supervised Instruction Tuning for VQA tasks. 

Specifically, adopting the widely-used multi-scale adaptive cropping strategy~\cite{ye2023ureader}, the input image $\mathbf{X} \in \mathbb{R}^{H \times W \times 3}$ is initially divided into several non-overlapping sub-images $\{\mathbf{X}_{i} \in \mathbb{R}^{\iota \times \iota \times 3} | i \in \{1,2,..., N\}\}$. By default, $\iota$ is set to 448 and $N$ does not exceed 6. Additionally, the original image $\mathbf{X}$ is resized to a global image $\mathbf{X}_{g}$ with the same size to preserve the overall layout. Subsequently, our proposed TokenFD processes these images $\mathcal{X} = \{\mathbf{X}_{g}, \mathbf{X}_{1}, ..., \mathbf{X}_{N} \}$ to produce their corresponding visual embeddings, denoted as $\mathcal{F} = \{\mathbf{\tilde{F}}_{i}\in \mathbb{R}^{\frac{4\times \iota}{p}\times \frac{4\times \iota}{p}\times D} | i \in \{g, 1,2,..., N\}\}$. 

After that, for each visual image features $\mathbf{\tilde{F}}_{i}$ (global image and sub-images), we apply a token abstractor $\xi: \mathbb{R}^{\frac{4\times \iota}{p}\times \frac{4\times \iota}{p}\times D} \xrightarrow{} \mathbb{R}^{\frac{\iota}{p \times \frac{s}{4}}\times \frac{\iota}{p\times \frac{s}{4}} \times D}$ to adaptively extract a meaningful visual embedding within each window of shape $s\times s$,
where $s$ is set to 4 in our experiment.
Specifically, in addition to the original dictionary of the tokenizer, we define a special token \texttt{<text>} to obtain a learnable token embedding $\mathbf{e}_{s} \in \mathbb{R}^{1\times 1\times D}$. Benefiting from the priors of the TokenFD, the special token embedding can easily learn robust representations to identify the most suitable visual embeddings within each window. Concretely, for each sub-image and global image, we first re-organize the shape of its visual embeddings $\mathbf{\tilde{F}}_{i}$ from $\frac{4\times \iota}{p}\times \frac{4\times \iota}{p}\times D$ to $(\frac{\iota}{p \times \frac{s}{4}})^{2} \times D \times s^{2}$. $\xi(\cdot)$ is then implemented as follows:
% {\setlength\abovedisplayskip{1pt}
% \setlength\belowdisplayskip{1pt}
\begin{equation}
  \begin{cases}
    \begin{aligned}
        \alpha_{i} &= \texttt{softmax} (\mathbf{e}_{s}\mathbf{\tilde{F}}_{i}), \alpha_{i}\in \mathbb{R}^{(\frac{\iota}{p \times \frac{s}{4}})^{2}\times 1\times s^{2}}\\
        \mathbf{\mathring{F}}_{i} &= \texttt{sum}(\alpha_{i} \circ \mathbf{\tilde{F}}_{i}), \mathbf{\mathring{F}}_{i} \in \mathbb{R}^{\frac{\iota}{p \times \frac{s}{4}}\times \frac{\iota}{p\times \frac{s}{4}} \times D}\\
    \end{aligned}
    \end{cases}
\end{equation}
% }
where the \texttt{softmax} and \texttt{sum} operations are conducted on the last dimension. $\circ$ denotes the Hadamard product. After the token abstractor, we flatten these compressed features $\{\mathbf{\mathring{F}}_{g},\mathbf{\mathring{F}}_{1},...,\mathbf{\mathring{F}}_{N}\}$ to get the final visual embeddings $\mathcal{V} = \{\mathbf{v}_{1},...,\mathbf{v}_{n_{v}}\}$, which will be fed into LLM.
Here, $n_{v}=\frac{\iota}{p \times \frac{s}{4}}\times \frac{\iota}{p\times \frac{s}{4}} \times (N+1)$ denotes the number of image tokens. 

% \subsubsection{LLM-guided Token Alignment}
\noindent \emph{\textbf{1) LLM-guided Token Alignment Training.}}

\begin{figure}[t]
    \centering
    \includegraphics[width=0.9\linewidth]{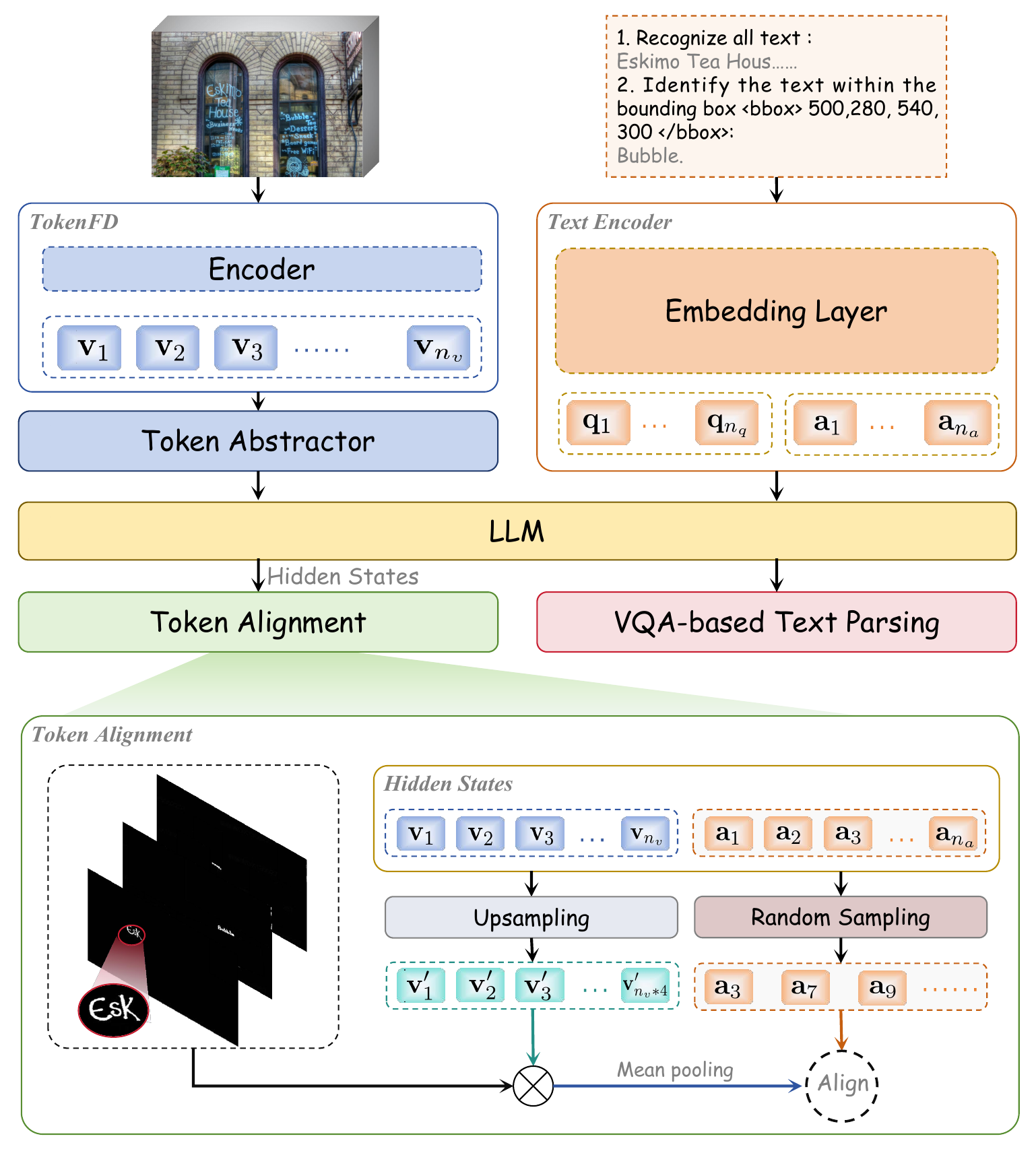}
    % \vspace{-0.5em}
    \caption{The framework of LLM-guided Token Alignment Training. Existing MLLMs primarily enhance spatial-wise text perception capabilities by integrating localization prompts to predict coordinates. However, this implicit method makes it difficult for these models to have a precise understanding. In contrast, the proposed token alignment uses BPE token masks to directly and explicitly align text with corresponding pixels in the input image, enhancing the MLLM's localization awareness. 
    % Thus, the same method can be easily integrated in the training stage of existing MLLMs and then discard the mask part and reuse the existing inference pipeline.
    }
    \label{fig:LLM}
    % \vspace{-0.5em}
\end{figure}

In the pre-training stage, we use the compressed visual embeddings $\mathcal{V}$ as the visual inputs, and $\mathcal{Q}$ and $\mathcal{A}$ from the Eq.\ref{eq:sample} as the language inputs to simultaneously conduct VQA-based text parsing tasks (\emph{implicitly semantic alignment}) and token alignment (\emph{explicitly spatial alignment}) tasks, as illustrated in Figure \ref{fig:LLM}. It is important to note that the Token Alignment (TA) branch is just introduced during LLM-guided Token Alignment Training, as all answers appear directly in the image. 

Text parsing tasks include recognizing full text, recognizing partial text within localization, visual text grounding, converting formulas into LaTeX, converting tables into markdown or LaTeX, and converting charts into CSV or markdown formats. More specific details are introduced in Supplementary Materials. Concretely, the visual and language inputs are concatenated together to be fed into the LLM, which predicts answers step-by-step by $\texttt{LLM}([\mathcal{V}_{1:n_{v}};\mathcal{Q}_{1:n_{q}};\mathcal{A}_{1:m-1}]),\forall m \in \{2,...,n_{a}\}$. The cross-entropy loss is formulated as:
% using the following optimization objective
{\setlength\abovedisplayskip{2pt}
\setlength\belowdisplayskip{2pt}
\begin{equation}
         \mathcal{L}_{cel} = - \sum_{m=2}^{n_{a}} \sum_{i=1}^{N}\mathbf{a}_{m}\log \hat{\mathbf{a}}_{m},\\
\end{equation}
}
where $\hat{\mathbf{a}}_{m} \in \mathbb{R}^{Z}$ refers to the probability distribution, $\mathbf{a}_{m}$ is the one-hot vector of $a_{m}$, and $Z$ denotes the dictionary size of the tokenizer.

The auto-regressive training task above allows only language inputs to implicitly interact with visual inputs. Without explicitly spatially-aware supervision, the outputs may depend more on the LLM's robust semantic context capabilities rather than the VFM's image feature representations. To explicitly facilitate spatial-wise visual-language alignment at the LLM level, we conduct a fine-grained alignment task with token granularity by leveraging the BPE token-mask pairs $\{(e_{1}, \mathbf{M}_{1}), (e_{2}, \mathbf{M}_{2}), ..., (e_{n_{e}}, \mathbf{M}_{n_{e}}) \}$. Specifically, given that the outputs of the $k$-th hidden layer of the LLM as $\{ \mathcal{V}^{k}, \mathcal{Q}^{k}, \mathcal{A}^{k}\}$ = $\{\mathbf{v}_{1}^{k},...,\mathbf{v}_{n_{v}}^{k},\mathbf{q}_{1}^{k},...,\mathbf{q}_{n_{q}}^{k},\mathbf{a}_{1}^{k},...,\mathbf{a}_{n_{a}}^{k}\}$, we extract the visual features and language features corresponding to each BPE token. Taking the BPE token $e_{i}$ as an example, we first compute its index location in $\{ \mathcal{V}^{k}, \mathcal{Q}^{k}, \mathcal{A}^{k}\}$ as $|\mathcal{V}^{k}| + |\mathcal{Q}^{k}| + \zeta(e_{i}, \mathcal{A})$, where $\zeta(e_{i}, \mathcal{A})$ finds the position of $e_{i}$ in $\mathcal{A}$ according to the relation $e_{i} \in \mathcal{E}$ and $\mathcal{E} \in \mathcal{A}$. For easy reference, the position has been recorded in our JSON file, which corresponds to the value for the keyword \texttt{index\_in\_text}. Consequently, the selected language features can be easily obtained through indexing operations. 
Then, to extract the selected visual features corresponding to the BPE token $e_{i}$, we exclude the global visual features (global image) and reorganize the remaining visual features (all sub-images) in $\mathcal{V}^{k}$ to recover a complete feature map, denoted as $\mathbf{F}^{k}$. A mean-pooling operation yields the associated token-level visual features:
{\setlength\abovedisplayskip{1pt}
\setlength\belowdisplayskip{1pt}
\begin{equation}
    \begin{aligned}
    \texttt{average}(\mathbf{M}_{i}\circ\texttt{BI}(\mathbf{F}^{k})),\\
    \end{aligned}
\end{equation}
}
where $\texttt{BI}(\cdot)$ refers to the bilinear interpolation operation to match the feature resolution of $\mathbf{M}_{i}$. \texttt{average} means performing a global pooling operation on the features.
Finally, the visual-language modality at the token level is aligned by minimizing the objectives following Eq.\ref{eq:alignment}.

Building on this, we assist the LLM in achieving fine-grained semantic perception for document understanding. This enables the visual semantics of each image patch with text to be consistent with the language semantics of its corresponding BPE token at the LLM level.

\noindent \emph{\textbf{2) Supervised Instruction Tuning.}}

Following the final stage of the previous MLLMs, we collect the existing VQA datasets to conduct supervised instruction tuning. These datasets cover a wide range of scenarios, including Documents (DocVQA, InfoVQA, DeepForm, KLC, DocMatix, AI2D, KIE, DocReason25K), Tables (TabFact, WTQ, TableBench, TabMWP, TableVQA), Charts (ChartQA, FigureQA, DVQA, PlotQA, UniChart, GeoQA+, Sujet-Finance), Formulas (UniMER, HME100k), and Scene Texts (TextVQA, ST-VQA, OCR-VQA, IAM, EST-VQA, SynthDoG). 

During the Supervised Instruction Tuning stage, we cancel the token alignment branch as answers may not appear in the image for some reasoning tasks (\emph{e.g.,} \textit{How much taller is the red bar compared to the green bar?}). This also ensures no computational overhead during inference to improve the document understanding capability. Finally, we inherit the remaining weights from the LLM-guided Token Alignment and unfreeze all parameters to facilitate comprehensive parameter updates.

\section{Experiments}
% \subsection{Implementation Details}
\noindent \textbf{Implementation Details.}
To pre-train the TokenFD foundation model, we employ the AdamW optimizer alongside a cosine learning rate schedule, with a base learning rate set at 5e-4. The model undergoes pre-training for two epochs on the TokenIT dataset. For TokenVL, we leverage the well-trained TokenFD as the visual foundation model and InternLM~\cite{cai2024internlm2} as the language model.
Specifically, during the LLM-guided token alignment stage, InternLM remains frozen while we train the TokenFD and newly introduced token abstractor. This stage involves training for one epoch on the TokenIT dataset, utilizing a base learning rate of 2e-4. In the subsequent supervised instruction tuning stage, all parameters are fully trainable, with a base learning rate of 1e-5. These experiments are executed on 64 H800 GPUs. Additional implementation details about other downstream experiments are provided within each respective subtask and \textbf{Supplementary Material}.

\subsection{Effectiveness of TokenFD}
Our work focuses on developing a high-performing dataset-agnostic foundation model. Fine-tuning, because it adapts representations to each dataset during the fine-tuning phase, can compensate for and potentially mask failures to learn general and robust representations. As a result, employing zero-shot transfer or fitting a linear classifier on representations extracted from the model, and then measuring its performance across various datasets, is a common approach~\cite{CLIP}. This method provides a clearer assessment of VFMs' ability to generalize without relying on dataset-specific tuning. 

\begin{table}[t]
    \centering
    \scalebox{0.7}{
    \begin{tabular}{l|l|c|C{8mm}C{10mm}C{12mm}|c}
     \toprule
     Tasks &Method & \#Param & TextSeg & TotalText & HierText &average\\
     \midrule
     \multirow{5}{*}{ZS} & CLIP-L-336px &304M & 19.71 & 13.56 & 13.39 &15.55\\
     &CLIP-L-448px &304M& 20.50 & 13.91 & 13.19&15.86\\
     &CLIP-L-1024px  &304M& 21.35 & 14.33 & 11.77&15.81\\
     % \cline{2-7}
     \rule{0pt}{2.4ex} %\cellcolor{mygray}
     &\cellcolor{mygray}TokenFD-448px  &\cellcolor{mygray}323M& \cellcolor{mygray}38.27 & \cellcolor{mygray}33.10 & \cellcolor{mygray}26.46&\cellcolor{mygray}32.61\\
     &\cellcolor{mygray}TokenFD-1024px  &\cellcolor{mygray}323M&\cellcolor{mygray} \textbf{38.28} &\cellcolor{mygray} \textbf{33.54} &\cellcolor{mygray} \textbf{31.95}&\cellcolor{mygray}\textbf{34.59}\\
     \midrule
     \multirow{3}{*}{LP}&SAM-H  &632M& 40.82 & 36.83 & 25.87&34.51\\
     &InternViT2.5  &300M &49.77 &42.54 &34.31&42.21\\
     % \cdashline{2-7}[4pt/2pt]
     \rule{0pt}{2.4ex} 
     &\cellcolor{mygray}TokenFD &\cellcolor{mygray}323M &\cellcolor{mygray}\textbf{55.66}  &\cellcolor{mygray}\textbf{47.53}  &\cellcolor{mygray}\textbf{43.11} &\cellcolor{mygray}\textbf{48.77}\\
     \bottomrule
\end{tabular}

    }
    \vspace{-0.5em}
    \caption{Text segmentation experiments of various visual foundation models. ``ZS'' refers to the zero-shot experiment. ``LP'' denotes the linear probe experiment.}
    \label{tab:segmentation}
    \vspace{-0.5em}
\end{table}

\noindent \textbf{Text Segmentation:} 1) Zero-shot Segmentation: 
We compute the similarity between visual and language features to get the segmentation results. For CLIP, in line with prior work, we select ``text'' as the language prompt, which has been proven to be the most effective~\cite{TCM}. In our method, we use a space `` '' as the language prompt and then apply a negation operation to derive the foreground similarity map.
2) Linear Probe: We keep the VFM frozen and train a linear layer to perform segmentation. 
Based on the results shown in Table \ref{tab:segmentation}, TokenFD demonstrates significant average performance improvement across various text segmentation tasks. In the zero-shot setting, TokenFD-1024px achieves the highest average score of 34.59\%, significantly outperforming CLIP-L by 18.78\%. In the linear probe setting, TokenFD again leads with an average score of 48.77\%, showing considerable improvements over SAM-H and InternViT2.5. 

\begin{table}[t]
    \centering
    \scalebox{0.7}{
    \begin{tabular}{l|c|C{10mm}C{10mm}C{10mm}C{14mm}|c}
\toprule
Method & \#Param  &  DocVQA & InfoVQA & TextVQA & ChartQA &average \\
\midrule
SAM-H & 632M &17.0 &23.1 &33.1 &30.1 &25.82\\
CLIP-L & 304M &64.9 &38.6 &80.7 &65.2 &62.36 \\
InternViT2.5 &300M&77.3 &49.3 &84.4 &74.0 &71.25\\
\midrule
\cellcolor{mygray}TokenFD &\cellcolor{mygray}323M &\cellcolor{mygray}\textbf{78.9} &\cellcolor{mygray}\textbf{50.0} &\cellcolor{mygray}\textbf{85.6} &\cellcolor{mygray}\textbf{74.4} &\cellcolor{mygray}\textbf{72.21}\\
\bottomrule
\end{tabular}
    }
    \vspace{-0.5em}
    \caption{The ANLS results of various visual foundation models on VQA tasks.}
    \label{tab:ZS_VQA_1}
    \vspace{-1.0em}
\end{table}
\noindent \textbf{Visual Question Answering:}
To further explore the representation learning capabilities of VFMs, we keep them frozen and fine-tune Vicuna-7B~\cite{zheng2023judging} as the language model to conduct the text-related VQA tasks. All comparison methods employ the same configuration—training data, test benchmarks, learnable parameters, and optimizer—to ensure a fair evaluation. 
As seen in Table \ref{tab:ZS_VQA_1}, TokenFD achieves the highest scores on popular benchmarks, outperforming SAM-H, CLIP-L, and InternViT2.5 by 46.39\%, 9.85\%, and 0.96\%, respectively. 

\begin{table}[t]
    \centering
    \scalebox{0.7}{
    \begin{tabular}{c|c|c|cc|c}
    \toprule
    Tasks & Methods &\#Param  & CTR (EN) & CSVTRv2 (CH) & average\\
    \midrule
     \multirow{3}{*}{LP} &CLIP-L &304M   &1.21 &6.03 &3.62 \\
     &InternViT2.5&300M &4.21 &22.37 &13.29 \\
     \rule{0pt}{2.4ex} 
    &\cellcolor{mygray}TokenFD &\cellcolor{mygray}323M &\cellcolor{mygray}\textbf{43.04} &\cellcolor{mygray}\textbf{84.19} &\cellcolor{mygray}\textbf{63.62} \\
    \bottomrule
\end{tabular}
    }
    \vspace{-0.5em}
    \caption{Linear probe experiments of various VFMs on text retrieval tasks. All VFMs are frozen.}
    \label{tab:STR}
    \vspace{-1.5em}
\end{table}

\noindent \textbf{Text Retrieval:}
We select representative models, CLIP and InternViT2.5, to compare with our proposed TokenFD on a Chinese dataset and an English dataset. Specifically, all VFMs are frozen. We calculate the similarity maps between the visual embeddings (extracted from the VFM) of all retrieval images and the language embeddings of all queries. 
For linear probe experiments, we use the same training data and train a simple linear classifier to score each similarity map, assigning a 1 if the similarity score is greater than 0.5, and a 0 otherwise. Finally, mean Average Precision (mAP) is employed to evaluate the performance of each VFM. The comparison results show that using only a few parameters, TokenFD can perform well. Specifically, our proposed TokenFD can achieve an average score of 63.62\% on bilingual tasks. Additionally, there remains significant room for improvement through specific designs and components in the future.

\vspace{-0.5em}
\subsection{Effectiveness of TokenVL}
\vspace{-0.8em}
\begin{table*}[t]
    \centering
    \begin{minipage}{\textwidth}
    \centering
    \scalebox{0.8}{
    \begin{tabular}{c|ccccccccccc}
        \toprule
        \cellcolor{mygray}\textbf{8B-Model} &ShareGPT4V  &Cambrian  & MM1.5 &POINT1.5 &GPT-4o &Gemini-1.5-Pro & GLM-4v  &Claude3.5 & InternVL2.5 \\
        \midrule
        \cellcolor{mygray}Score &398  &614   & 635 &720 &736 & 754 & 776 &788 & 822 \\
        \bottomrule
    \end{tabular}
    }
    \end{minipage}
    \vspace{0.5em}
    \begin{minipage}{\textwidth}
    \centering
    \scalebox{0.8}{
    \begin{tabular}{c|cccc|c|cccccc}
        \toprule
        8B-Model & TextMonkey & DocOwl-1.5 & TextHawk2 & \cellcolor{mygray}TokenVL(ours) & \cellcolor{mygray}\textbf{2B-Model} & MiniMonkey & InternVL2.5 & \cellcolor{mygray}TokenVL(ours)\\
        \midrule
        Score & 561 & 599 & 784 & \cellcolor{mygray}\textbf{860} &\cellcolor{mygray}Score  & 802 &804 &\cellcolor{mygray}\textbf{821} \\
        \bottomrule
    \end{tabular}
    }
    \end{minipage}
    \vspace{-1.5em}
    \caption{Comparison results of our TokenVL with other MLLMs on the OCRbench benchmark.}
    \vspace{-0.5em}
    \label{tab:OCRbench}
\end{table*}
\vspace{1em}
\begin{table*}[t]
    \centering
    \scalebox{0.73}{
    
\begin{tabular}{l|c|l|ccc|c|c|cc|ccc}     
        \toprule 
        Model  & size &Venue & DocVQA & InfoVQA & DeepForm & ChartQA & TextVQA\textsubscript{Val} & WTQ & TabFact & FUNSD & SROIE  & KLC \\
        \midrule
        MiniCPM-V   & 3B &COLM'24 &71.9&- &-&55.6&74.1&-&-&-&-\\
        Mini-Monkey & 2B &ICLR'25 & 87.4 & 60.1 & - & 76.5 & 75.7 & - & - & 42.9 & 70.3 & -  \\
        InternVL2.5 & 2B &arxiv'24 &88.7&60.9&15.2&79.2&74.3&38.7&58.1&37.9&68.1&16.1\\
        \midrule
        TokenVL &2B&-&\textbf{89.9}&\textbf{61.0}&\textbf{71.9}&\textbf{81.1}&\textbf{76.4}&\textbf{49.0}&\textbf{76.9}&\textbf{43.0}&\textbf{82.6}&\textbf{38.8}\\
        \bottomrule
        \toprule
        Claude-3.5 Sonnet & \multicolumn{2}{c|}{Closed-source model}  & 88.5 & 59.1 & 31.4 & 51.8 & 71.4 & 47.1 & 53.5 &- &- &24.8\\
        GeminiPro-1.5 & \multicolumn{2}{c|}{Closed-source model} & 91.2 & 73.9 & 32.2 & 34.7 & 80.4 & 50.3 & 71.2 &- &- &24.1\\
        GPT4o 20240806 &\multicolumn{2}{c|}{Closed-source model} &92.8&66.4&38.4&85.7&70.5&46.6&81.1&-&-&29.9\\
        \midrule
        DocPeida & 7B &arxiv'23 & 47.1 & 15.2 & - & 46.9 & 60.2 & - & - & 29.9 & 21.4 & -   \\
        DocOwl & 7B &arxiv'23 & 62.2 & 38.2 & 42.6 & 57.4 & 52.6 & 26.9 & 67.6 & 0.5 & 1.7 & 30.3   \\
        LLaVA1.5 & 7B & NeurIPS'23 & - & - & - & 9.3 & - & - & - & 0.2 & 1.7 & - \\
        UReader & 7B &EMNLP'23 & 65.4 & 42.2 & 49.5 & 59.3 & 57.6 & 29.4 & 67.6 & - & - & 32.8   \\ 
        % Dessurt & <1B & 63.2 & - & - & - & - & - & - & - & - & - & -   \\         
        % Donut & <1B & 67.5 & 11.6 & 61.6 & 30.0 & 41.8 & 43.5 & 18.8 & 54.6 & - & - & -   \\ 
        % Pix2Struct & 1.3B & 76.6 & 40.0 & - & - & 58.6 & - & - & - & - & - & -   \\ 
        % QwenVL~\cite{bai2023qwen} & 9.6B &arxiv'23 & 65.1 & 35.4 & - & - & 65.7 & 63.8 & - & - & - & - & -   \\ 
        % VisCoT-7B & 47.6 & - & - & - & - & \textbf{77.5} & - & - & - & 47.0 & -   \\
        CHOPINLLM & 7B &arxiv'24 & - & - & - & 70.0 & - & - & - & - & - & -   \\ 
        TextHawk & 7B &arxiv'24 & 76.4 & 50.6 & -  & 66.6 & - & 34.7 & 71.1 & - & - & -   \\ 
        DocKylin & 7B &arxiv'24 & 77.3 & 46.6 & -  & 66.8 & - & 32.4 & - & - & - & -   \\ 
        MM1.5 & 7B &arxiv'24 & 88.1 & 59.5 & -  & 78.6 & 76.8 & 46.0 & 75.9 & - & - & -   \\ 
        DocOwl-1.5 & 8B &EMNLP'24 & 81.6 & 50.4 & 68.8  & 70.5 & 68.8 & 39.8 & 80.4 & - & - & 37.9   \\
        DocOwl-1.5-Chat & 8B &EMNLP'24 & 82.2 & 50.7 & 68.8  & 70.2 & 68.6 & 40.6 & 80.2 & - & - & 38.7  \\
        CogAgent & 17B &CVPR'24 & 81.6 & 44.5 & -  & 68.4 & 76.1 & - & - & - & - & -   \\
        Monkey & 10B &CVPR'24 & 66.5 & 36.1 & 40.6  & 65.1 & 67.6 & 25.3 & - & - & - & -   \\
        TextMonkey & 8B &arxiv'24 & 73.0 & 28.6 & -  & 66.9 & 65.6 & - & - & 32.3 & 47.0 & - \\
        HRVDA & 7B & CVPR'24 & 72.1 & 43.5 & 63.2  & 67.6 & 73.3 & 31.2 & 72.3 & - & - & 37.5 \\
        InternVL2 & 8B & CVPR'24 & 91.6 & 74.8 & -  & - & 
        77.4 & - & - & - & - & -    \\
        Park et al. & 7B & NeurIPS'24 & 72.7 & 45.9 & 53.0 & 36.7 & 59.2 & 34.5 & 68.2 & - & - & 36.7   \\
        MOAI & 7B &ECCV'24 & - & - & - & - & 67.8 & - & - & - & - & -   \\ 
        Vary & 7B & ECCV'24 & 76.3 & - & - & 66.1 & - & - & - & - & - & - \\
        TextHawk2 & 7B &arxiv'24 & 89.6 & 67.8 & - & 81.4 & 75.1 & 46.2 & 78.1 & - & - & -   \\ 
        PDF-WuKong & 9B &arxiv'24 & 76.9 & - & - & - & - & - & - & - & - & -   \\
        % TCC & 8B & arxiv'24 & 78.3 & 50.2 & 65.7 & 68.9 & 66.6 & 38.6 & 79.3 & - & - & 35.9  \\
        % Marten(w/o MGM)$\dag$ & 8.1B & - & 89.5 & 71.7 & 71.2 & 36.4 & 79.3 & 71.3 & 49.5 & 81.6 & 41.1 & 78.7 & 66.9  \\
        % Marten & 8B & - & \textbf{92.0} & \textbf{75.2} & \textbf{75.1} & \textbf{39.5} & \underline{81.7} & 74.4 & \textbf{52.4} & \textbf{84.4} & \textbf{44.4} & \textbf{80.4} & \underline{69.5}   \\ 
        LLaVA-NEXT-7B & 7B & arxiv'24& 63.5 & 30.9 & 1.3 & 52.1 & 65.1 & 20.1 & 52.8 & - &- &5.35\\
        LLama3.2-11B & 11B & arxiv'24 & 82.7 & 36.6 & 1.78 & 23.8 & 54.3 & 23.0 & 58.3 & - &- &3.47\\
        Pixtral-12B & 12B & arxiv'24 & 87.7 & 49.5 & 27.4 & 71.8 & 76.1 & 45.2 & 73.5 & - &- & 24.1 \\
        Ovis & 9B & arxiv'24 & 88.8 & 74.0 & 45.2 & 81.4 & 77.7 & 50.7 & 76.7 &- &- & 23.9\\
        InternVL2.5 & 8B &arxiv'24&93.0&\textbf{77.6}&37.9&84.8&79.1&52.7&74.8&38.26&71.7&22.9 \\
        AlignVLM & 8B & arxiv'25 & 81.2 & 53.8 & 63.3 & 75.0 & 64.6 & 45.3 & 83.0 & - &- & 35.5 \\
        \midrule
        \cellcolor{mygray}TokenVL w/o TA &\cellcolor{mygray} 8B &\cellcolor{mygray}- &\cellcolor{mygray}\underline{93.8} &\cellcolor{mygray}75.3&\cellcolor{mygray}\underline{72.4}&\cellcolor{mygray}\underline{86.5}&\cellcolor{mygray}\underline{79.3}&\cellcolor{mygray}\underline{57.2}&\cellcolor{mygray}\underline{83.6}&\cellcolor{mygray}\underline{41.5}&\cellcolor{mygray}\underline{79.0}&\cellcolor{mygray}\underline{39.6} \\
        \cellcolor{mygray}TokenVL &\cellcolor{mygray} 8B &\cellcolor{mygray}-&\cellcolor{mygray}\textbf{94.2}&\cellcolor{mygray}\underline{76.5}&\cellcolor{mygray}\textbf{72.9}&\cellcolor{mygray}\textbf{86.6}&\cellcolor{mygray}\textbf{79.9}&\cellcolor{mygray}\textbf{61.4}&\cellcolor{mygray}\textbf{85.2}&\cellcolor{mygray}\textbf{42.2}&\cellcolor{mygray}\textbf{81.9}&\cellcolor{mygray}\textbf{39.9} \\
        \bottomrule
    \end{tabular}
    }
    \vspace{-0.9em}
    \caption{Comparisons on various types of text-rich image understanding tasks. All evaluation benchmarks use the officially designated metrics. ``size" refers to the number of parameters in the model, and ``Val" refers to the validation set.}
    \label{tab:VQA}
    \vspace{-0.5em}
\end{table*}

\noindent \textbf{OCRBench results:}
OCRBench is a widely recognized and comprehensive benchmark comprising 29 tasks, commonly utilized to assess the OCR capabilities of MLLMs. As illustrated in Table \ref{tab:OCRbench}, we compare the performance of our TokenVL against previously existing MLLMs. TokenVL achieves the highest score of 860 among the 8B-Model group, significantly outperforming models like general-MLLM InternVL2.5 ($\uparrow38$) and expert TextHawk2 ($\uparrow76$). In the 2B-Model group, our method achieves the top score of 821, surpassing competitors such as MiniMonkey ($\uparrow19$) and InternVL2.5 ($\uparrow17$).

\noindent \textbf{Document Benchmarks results:}
To demonstrate the perception, understanding, and reasoning capabilities of our TokenVL, we collect existing evaluation benchmarks across five categories: Document, Chart, Natural Scene, Table, and KIE. The results, presented in Table \ref{tab:VQA}, show a consistent and significant outperformance over other 8B MLLMs. Specifically, for widely used evaluation benchmarks (Doc/Info/Chart/TextVQA), TokenVL-2B achieves an average gain of 2.18\% and 1.33\% over MiniMonkey and InternVL2.5, respectively. TokenVL-8B obtains gains of 1.2\%, 1.8\%, and 0.8\% on DocVQA, ChartQA, and TextVQA compared to the previous SOTA InternVL2.5. Additionally, TokenVL achieves a larger performance gain on other benchmarks while maintaining these properties.

\subsection{Ablation Study}

\begin{table}[t]
    \centering
    \scalebox{0.73}{
    
\begin{tabular}{c|cccc}
    \toprule
     Method & TotalText ($\downarrow$) & IC15 ($\downarrow$) & IIT ($\downarrow$) & Docgenome ($\downarrow$) \\
     \midrule
    %  InternVL2-5 & & & & \\
     w/o token alignment &0.3592 &0.2388 &0.2388 &0.2837 \\
     w token alignment &\textbf{0.3547} &\textbf{0.2324} &\textbf{0.1921} &\textbf{0.2806} \\
     \bottomrule
\end{tabular}
    }
    \vspace{-0.5em}
    \caption{Edit distance for full-image text recognition.}
    \label{tab:NPTS}
    \vspace{-0.9em}
\end{table}
\begin{table}[]
    \centering
    \scalebox{0.73}{
    \begin{tabular}{C{12mm}C{13mm}|C{10mm}C{11mm}C{13mm}C{14mm}}
    \toprule
    Abstractor & Alignment & DocVQA & InfoVQA & ChartVQA & TextVQA\textsubscript{Val} \\
     \midrule
     $\times$ & $\times$ &93.1 &74.7 &86.5 &79.1\\
     \checkmark & $\times$ &93.8 &75.3  &86.5  &79.3\\
     \checkmark &\checkmark &\textbf{94.2} &\textbf{76.5} &\textbf{86.6} &\textbf{79.9} \\
     \bottomrule
\end{tabular}
    }
    \vspace{-0.6em}
    \caption{Comparison experiments on the VQA tasks.}
    \label{tab:four_VQA}
    \vspace{-1.0em}
\end{table}

\noindent \textbf{w/o token alignment.}
Token alignment at the LLM level explicitly facilitates interaction between image embeddings and language embeddings. This method encourages the LLM to reference image content more directly when responding to questions, rather than relying solely on its powerful semantic context capabilities. To verify the effectiveness of this strategy:
1) we perform a text recognition experiment of full-text images, which predicts all texts within a given image from top to bottom and left to right. 
As shown in Table \ref{tab:NPTS}, without fine-tuning on downstream text data, we directly evaluate our model's performance with and without Token Alignment, using document scenes (1000 images extracted from IIT-CDIP and DocGenome respectively) and natural scenes (ICDAR15 and TotalText). Specifically, given the question ``recognize all texts in the image" for MLLMs, we calculate edit distance by comparing the model's outputs with the ground truth answers sorted by spatial position. It was observed that token alignment significantly improves text recognition performance on full images.
2) we also evaluate the final VQA performance of the MLLM on four widely used evaluation benchmarks, both with and without Token Alignment, referring to the last two group results of Table \ref{tab:four_VQA}. As a result, an average gain of 0.6\% is obtained. More details are provided in Supplementary Material.

\noindent \textbf{w/o token abstractor.} To reduce the spatial dimensions, we designed a learnable token embedding vector to adaptively capture useful visual information. Without the token abstractor, we use a simple pooling layer instead. The ablation results are shown in the top two groups of Table \ref{tab:four_VQA}, where an average gain of 0.3\% is obtained, even though the token abstractor is not our main contribution.
\vspace{-0.1em}
\section{Conclusion}
In the paper, we take a step towards constructing a fine-grained visual foundation model, and propose a series of token-level product families: TokenIT, TokenFD, and TokenVL. We also explore the potential and effectiveness of TokenFD and TokenVL at a sufficiently large scale. While this approach demonstrates good and consistent performance gains on downstream tasks, there remains significant room for improvement through effective training strategies or additional designs. Therefore, we hope these products will serve as easily reproducible baselines for more downstream tasks in the future.

\nocite{langley00}

\bibliography{example_paper}
\bibliographystyle{icml2025}

%%%%%%%%%%%%%%%%%%%%%%%%%%%%%%%%%%%%%%%%%%%%%%%%%%%%%%%%%%%%%%%%%%%%%%%%%%%%%%%
%%%%%%%%%%%%%%%%%%%%%%%%%%%%%%%%%%%%%%%%%%%%%%%%%%%%%%%%%%%%%%%%%%%%%%%%%%%%%%%
% APPENDIX
%%%%%%%%%%%%%%%%%%%%%%%%%%%%%%%%%%%%%%%%%%%%%%%%%%%%%%%%%%%%%%%%%%%%%%%%%%%%%%%
%%%%%%%%%%%%%%%%%%%%%%%%%%%%%%%%%%%%%%%%%%%%%%%%%%%%%%%%%%%%%%%%%%%%%%%%%%%%%%%
\newpage
\appendix
\onecolumn

\section{VQA-based Text Parsing Tasks}
Modality connectors act as the bridge between the visual foundation model (VFM) and the LLM. Previous MLLMs employ image-text pairs of natural images (\emph{e.g.,} Conceptual Captions, LAION, COYO) to pre-train them. In the work, to endow our MLLM TokenVL with generality and comprehensive document understanding abilities, we follow DocOwl~\cite{mplug-docowl1.5} to conduct modality alignment. It involves both structure-aware parsing tasks (recognizing full text, converting formulas into LaTeX, converting tables into markdown or LaTeX, and converting charts into CSV or markdown formats) and multi-grained text localization tasks (recognizing partial text within localization, visual text grounding). Specifically, we present an example to introduce them, as shown in Table~\ref{tab:TextParsing}. In this way, the pre-trained modality connector can understand the visual features of VFM and project them into the same feature space with the language features of LLM. 

\begin{table}[htbp]
    \centering
    \scalebox{0.9}{
    \begin{tabular}{p{0.15\linewidth}p{0.75\linewidth}}
\toprule
\multicolumn{2}{l}{\textbf{VQA-based Text Parsing}}  \\
\midrule

\multicolumn{2}{c}{\includegraphics[width=0.95\linewidth]{Figure/chart0.png} } \\
\midrule
Question: & \textcolor{blue}{Recognizing full text.} \\[0.4em]
Answer: & CO${}_2$ emissions by mode of transport, 2018.\\
 &Carbon dioxide (CO2) emissions are measured in grams per passenger kilometer. This does not account for non-CO2 greenhouse gases or the $\cdots\cdots$.\\
\midrule
Question: & \textcolor{blue}{Recognizing the text within the bounding box $<$bbox$>$75, 200, 160, 230$<$/bbox$>$.}\\[0.4em]  
Answer: & Medium car (diesel). \\
\midrule
Question: & \textcolor{blue}{Pblueict the bounding box of the text $<$ocr$>$Eurostar (International rall)$<$/ocr$>$ } \\[0.4em]
Answer: & $<$bbox$>$25, 520, 160, 550$<$/bbox$>$. \\
\midrule
Question: & \textcolor{blue}{Converting the chart into CSV format.} \\[0.4em]
Answer: & Mode of Transport, CO2 Emissions (g CO2 per passenger km)\\
 &Medium car (petrol), 191.6\\
 &Medium car (diesel), 168.8\\
 &Domestic flight, 133.5\\
 &Bus, 103.9\\
 &$\cdots\cdots$.\\
\midrule
Question: & \textcolor{blue}{Converting the chart into Markdown format.} \\[0.4em]
Answer: & \textbar Mode of Transport \textbar CO2 Emissions (g CO2 per passenger km) \textbar \\
 &\textbar---------------------\textbar---------------------\textbar\\
 &\textbar Medium car (petrol)             \textbar 191.6                                  \textbar\\
 &\textbar Medium car (diesel)             \textbar 168.8                                  \textbar\\
 &\textbar Domestic flight                  \textbar 133.5                                  \textbar\\
 &\textbar Bus                              \textbar 103.9                                  \textbar\\
 &$\cdots\cdots$.\\
\bottomrule
\end{tabular}

    }
    % \vspace{-0.5em}
    \caption{The illustration of VQA-based Text Parsing tasks of TokenVL.}
    \label{tab:TextParsing}
\end{table}
\section{Interactive Demo}
\begin{figure*}
    \centering
    \includegraphics[width=0.8\linewidth]{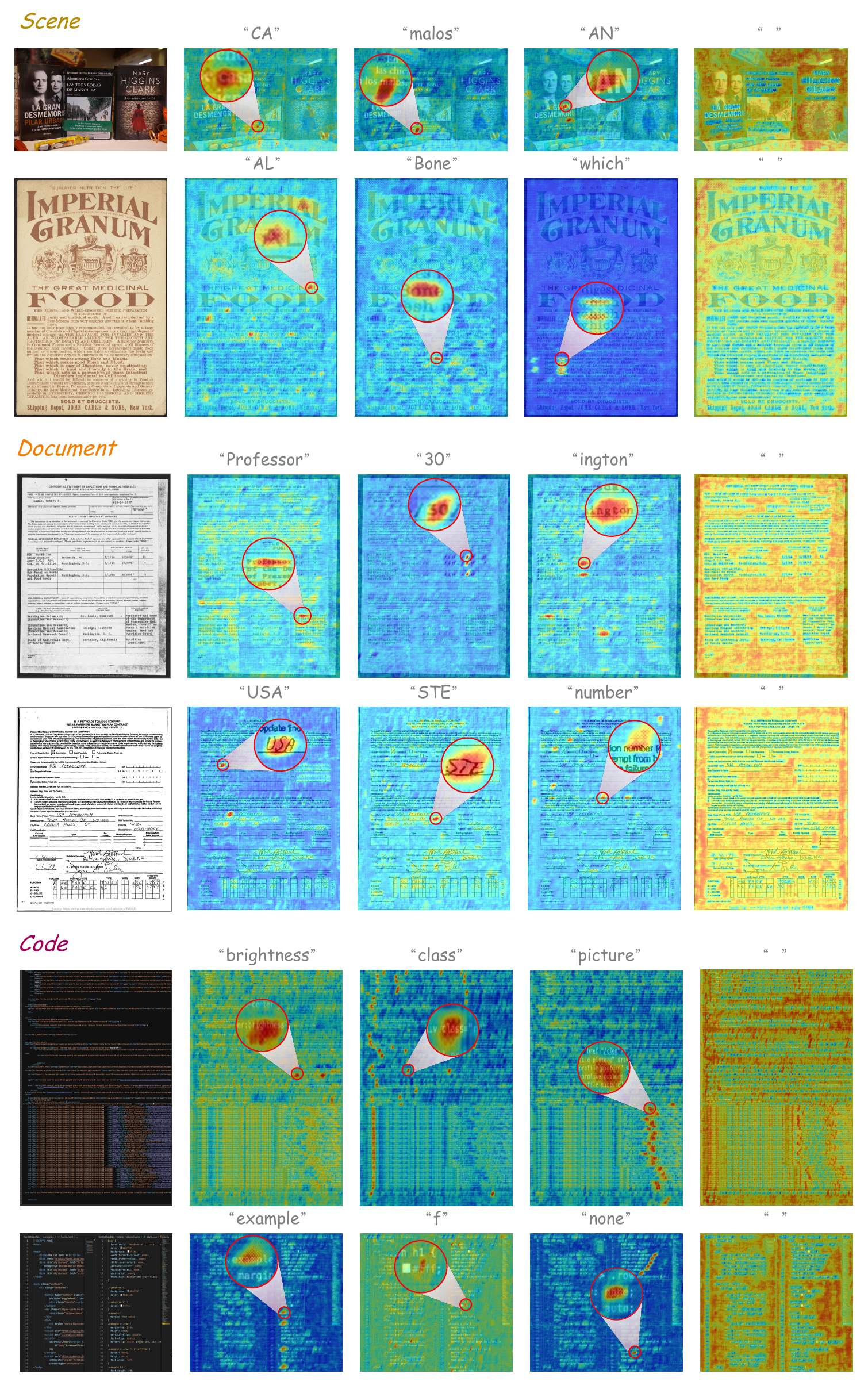}
    \caption{More visualization examples of the natural scene images, document images, and code images.}
    \label{fig:demo1}
\end{figure*}

\begin{figure*}
    \centering
    \includegraphics[width=0.8\linewidth]{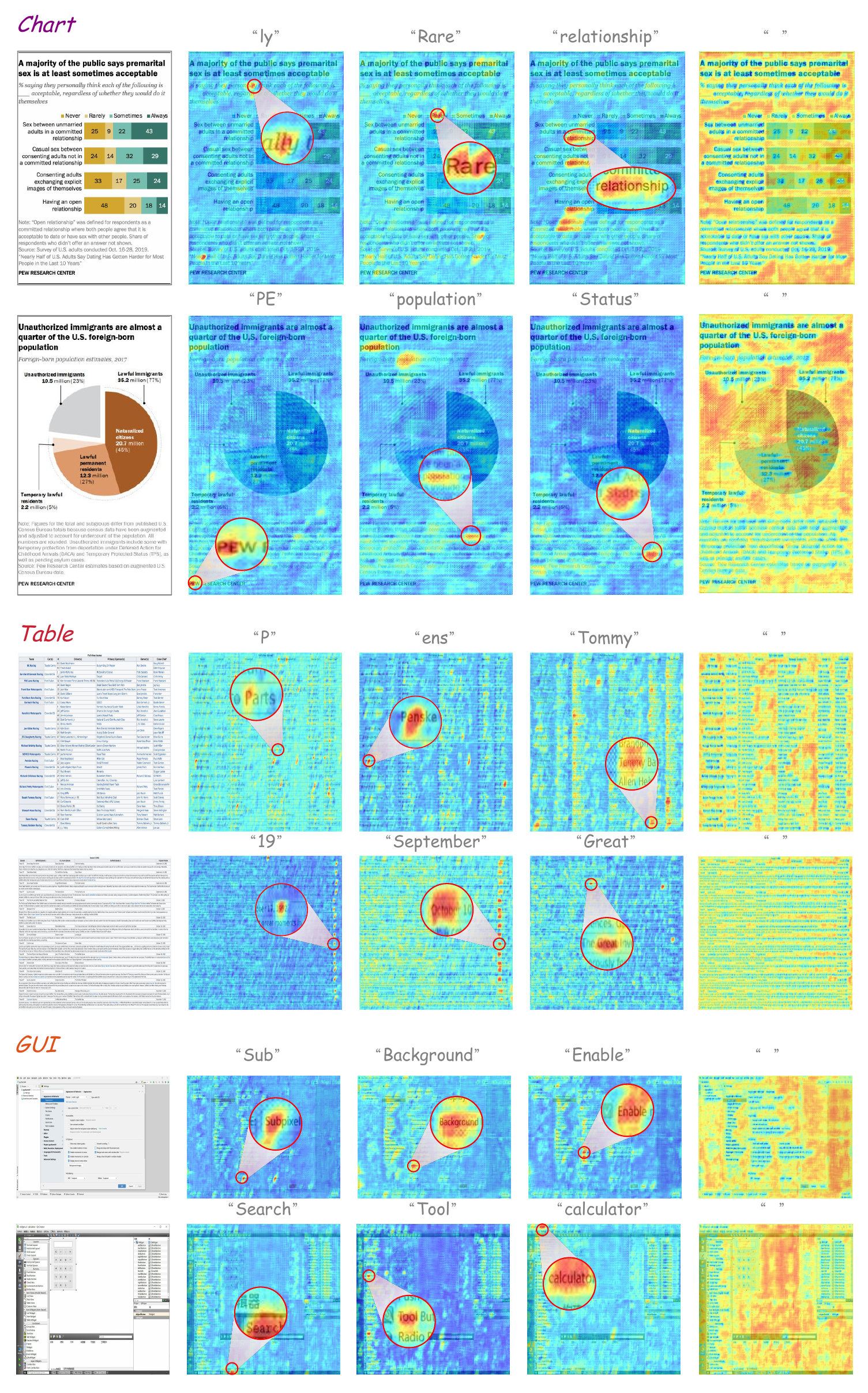}
    \caption{More visualization examples of the chart, table, and GUI images.}
    \label{fig:demo2}
\end{figure*}

As shown in Figure~\ref{fig:demo1} and Figure~\ref{fig:demo2}, we provide more interactive examples, including natural scene images, documents, codes, charts, tables, and GUIs. For each scene, we provide two examples. The first column is the original image, the second to fourth columns are visualizations of the corresponding BPE words within the image, and the last column shows the highlighted area of the image when the prompt is a space `` ''. As we observed, 

\textbf{1)} Our foundation model TokenFD can distinguish text and background areas well. This means that when using the foundation model for downstream tasks, we can remove redundant background features at very low cost; 

\textbf{2)} For complex, dense, and small texts, TokenFD still precisely perceive, such as ``picture''(Code), ``f''(Code), ``19''(Table), ``P''(Table), \emph{etc}. Additionally, our TokenFD still can capture punctuation marks, such as commas, periods, double quotes, \emph{etc}; 

\textbf{3)} Our TokenFD also supports handwritten texts, such as ``STE''(Document) and ``USA''(Document). We will provide a demo address for users to experience it.


\begin{figure*}
    \centering
    \includegraphics[width=0.8\linewidth]{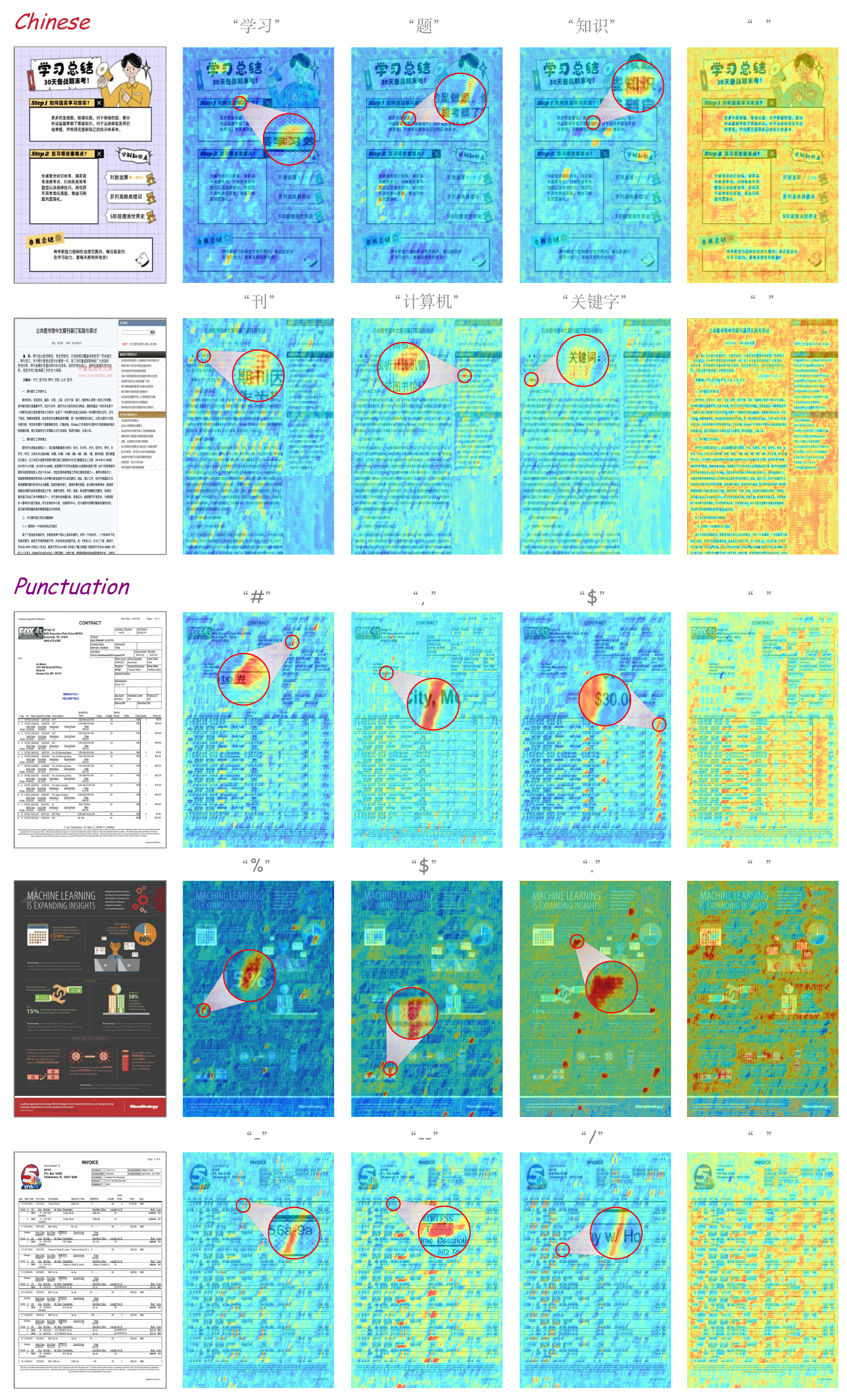}
    \caption{More visualization examples of the Chinese.}
    \label{fig:demo3}
\end{figure*}

\section{TokenIT Dataset}
\subsection{Data Source}
To construct a comprehensive TokenIT dataset, we collect various types of data, including natural scene text images, documents (PDF, receipt, letter, note, report, \emph{etc.}), tables, charts, and GUI. The data sources are summarized in Table~\ref{tab:TokenIT}. 
\begin{table}[htbp]
    \centering
    \begin{tabular}{c|p{12cm}}
    \toprule
    \textbf{Dataset Type} &  \textbf{Dataset Name} \\
    \midrule
    Natural Scene & ICDAR2013~\cite{karatzas2013icdar}, COCOText~\cite{veit2016coco}, CTW1500~\cite{yuliang2017detecting}, HierText~\cite{long2022towards}, ICDAR2015~\cite{cheng2017focusing}, OCRCC~\cite{kareem2017application}, OpenImagesV5Text~\cite{krylov2021open}, TextCaps~\cite{sidorov2020textcaps}, TextOCR~\cite{singh2021textocr}, TotalText~\cite{ch2017total}, Laion-OCR~\cite{schuhmann2021laion}, Wukong-OCR~\cite{gu2022wukong}, Mlt2017~\cite{nayef2017icdar2017}, ocrvqa~\cite{mishra2019ocr}, ST-VQA~\cite{biten2019scene}, SynText~\cite{liu2020abcnetrealtimescenetext}, the-cauldron~\cite{laurençon2024matters}, ArT~\cite{ArT}, ChineseOCR~\cite{ChineseOCR}, HCCDoc~\cite{hccdoc}, ICDAR2017rctw~\cite{rctw}, LSVT~\cite{LSVT}, MTWI~\cite{MTWI}, and ReCTS~\cite{ReCTS} \\
    \midrule
    Document & DocVQA~\cite{mathew2021docvqa}, InfographicsVQA~\cite{mathew2022infographicvqa}, KIE, KleisterCharity~\cite{stanislawek2021kleister}, PubTabNet~\cite{zhong2020image}, RVL-CDIP~\cite{harley2015icdar}, VisualMRC~\cite{tanaka2021visualmrc}, Docmatix~\cite{laurenccon2024building}, IIT-CDIP~\cite{Xu2019ARXIV_LayoutLM_Pre_training}, publaynet~\cite{zhong2019publaynet}, Synthdog-en~\cite{kim2022ocr}, DocGenome~\cite{xia2024docgenome}, CCpdf~\cite{turski2023ccpdf}\\
    \midrule
    Chart & ChartQA~\cite{masry2022chartqa}, FigureQA~\cite{kahou2017figureqa}, PlotQA~\cite{methani2020plotqa},TabMWP~\cite{lu2022dynamic}, DVQA~\cite{kafle2018dvqa}  \\
    \midrule
    Table & TableQA~\cite{sun2020tableqa},  DeepForm~\cite{svetlichnaya2020deepform}, TURL~\cite{deng2022turl}, TabFact~\cite{chen2019tabfact}, WikiTableQuestions~\cite{pasupat2015compositional} \\
    % \midrule
    % Code & Code\\
    \midrule
    GUI & Screen2Words~\cite{screen2words}, WebSight~\cite{websight}, OmniACT~\cite{omniact}, SeeCliCK~\cite{seeclick}, Mind2Web~\cite{mind2web}\\
    \bottomrule
\end{tabular}

    \caption{Data source of our TokenIT dataset.}
    \label{tab:TokenIT}
\end{table}

\subsection{Data Generation}
Next, we elaborate on the data construction pipeline for the TokenIT dataset, which involves four steps: 

\noindent \emph{\textbf{1) Text Image Segmentation.}} For natural scene text images, charts and tables, we fine-tune the SAM model~\cite{SAM} on datasets with character-level mask annotations and leverage the well-learned model to generate text masks, since these images are relatively complex and diverse in color and style. For PDFs and industrial documents, we conduct simple unsupervised clustering~\cite{kanungo2002efficient} to get their text masks;

\noindent \emph{\textbf{2) Text Recognition.}} We use the previous SOTA method~\cite{guan2023self} to obtain the recognition results for all types, except for natural scene text images. As these natural scene datasets already provide text transcriptions, we adopt them directly;  

\noindent \emph{\textbf{3) Tokenizer.}} We choose the widely adopted BPE tokenizer~\cite{internvl} to split the language texts into multiple BPE tokens, where each token corresponds to a BPE-level subword; 

\noindent \emph{\textbf{4) Token-level Image Text Construction.}} After obtaining the text masks in Step 1, we apply the method~\cite{guan2025ccdplus} to produce character-level segmentation masks. Subsequently, we combine each token's corresponding character-level masks to create a complete token-level segmentation mask.

\noindent \emph{\textbf{5) Data Correction.}}
For each image and its generated labels following the above stage, we render the labels onto the images to verify data labeling quality and perform manual relabeling as needed. Finally, \emph{three rounds of inspections are conducted to minimize labeling errors, a process that took four months to develop the first token-level image text dataset (TokenIT).}

Overall, the proposed TokenIT dataset includes 20 million images (including natural scene text images, documents, tables, charts, Code, and GUI) and 1780679833 (1.8 billion) token-mask pairs. Each BPE token corresponds one-to-one with a pixel-level mask. 
% The data ratios are summarized in Figure \ref{fig:dataset} (b). Figure \ref{fig:dataset} (c) and (d) further provide the number distribution of tokens per image type and a word cloud of the top 100 tokens, respectively.

% \subsection{Data Analysis}
% To construct a robust and comprehensive TokenIT dataset, we collect various types of data, including natural scene text images, documents (PDF, receipt, letter, note, report, \emph{etc.}), tables, charts, and GUI. The data ratios are summarized in Figure \ref{fig:dataset} (b). Overall, the proposed TokenIT dataset includes 20 million images and 1780679833 (1.8 billion) token-mask pairs. Each BPE token corresponds one-to-one with a pixel-level mask. Figure \ref{fig:dataset} (c) and (d) further provide the number distribution of tokens per image type and a word cloud of the top 100 tokens, respectively. 

\section{Training Details}
\subsection{Text Segmentation}
In this section, we evaluate the performance of text segmentation using TextSeg, COCOText, and HierText, which provide pixel-level annotations. The test sets of these datasets are utilized for zero-shot experiments. In the linear probe setting, all methods are trained on the combined three training sets and evaluated separately on each test set. The training configuration includes 70 epochs, a learning rate of 0.0001, a batch size of 6, and the optimizer AdamW. 

\subsection{Visual Question Answering}
In this section, we evaluate the performance of visual document understanding using the test sets of DocVQA, InfoVQA, ChartQA, and TextVQA. The proposed model undergoes a two-phase training process: pre-training and fine-tuning. 

During the pre-training phase, we randomly sampled 200,000 images each from the IIT-CDIP and DocMatix document datasets. Full-text recognition was implemented using PaddleOCR to generate ground-truth textual content, which served as target answers. The model was trained with the instructional prompt "Recognize all text:" where only the Multilayer Perceptron (MLP) component received parameter updates. The training configuration included one epoch with a learning rate of 1e-3 and a batch size of 24.

For the fine-tuning phase, we extended the training scope to incorporate Low-Rank Adaptation (LoRA) for Large Language Model (LLM) optimization while continuing MLP updates. The training data comprised the training splits of the aforementioned QA evaluation datasets. This phase maintained single-epoch training with modified hyperparameters, specifically a reduced learning rate of 2e-4 and a batch size of 12 to ensure stable parameter convergence. This hierarchical training paradigm progressively enhances both text recognition accuracy and semantic comprehension capabilities in document understanding tasks.

\subsection{Text Retrieval}
In this section, we evaluate model performance using the CTR benchmark (English)~\cite{veit2016coco} and the CSVTRv2 benchmark (Chinese)~\cite{CSVTR}. For English text retrieval, we employ the training sets from ICDAR2013, ICDAR2015, COCOText, MLT2017, OpenImagesV5Text, CTW1500, TotalText, HierText, and TextOCR. For Chinese text retrieval, we use ArT, ChineseOCR, HCCDoc, icdar2017rctw, LSVT, MTWI, and ReCTS as the training sets.
These methods are optimized using the AdamW optimizer. The initial learning rate is 1e-4. We use a batch size of 6 and a number of training epochs of 10. After the first 5 epochs, the initial learning rate is reduced to 1e-5.

\section{Mainstream Benchmark Results} 
General multi-modal large models typically use DocVQA, InfoVQA, ChartQA, and TextVQA to evaluate document understanding capabilities, as these benchmarks encompass diverse and comprehensive scenarios that reflect real-world applications. To compare performance intuitively and clearly, we collected data from nearly all models that reported scores on these four benchmarks and summarized them in Table~\ref{tab:popular}. Specifically, we categorized the existing MLLMs into three types based on model size: ``$<$2B'', ``$<$8B'', and ``$>$8B''. Due to resource constraints, we did not conduct experiments with models exceeding 8B parameters in our TokenVL, providing only two versions: TokenVL-2B and TokenVL-8B. Notably, our TokenVL-2B improves upon the previous state-of-the-art (SOTA) result by 1.32\%, and our TokenVL-8B improves by 0.63\%. Compared to models with larger parameters, our 8B version slightly surpasses DeepSeek-VL2-16B and InternVL2-40B by 0.3\%.

% \section{General Capability Experiment} 
% \subsection{Implementation Details}

\begin{table}[]
    \centering
        \centering
    \scriptsize
    \begin{tabular}{l|l|>{\centering\arraybackslash}m{3.5cm}>{\centering\arraybackslash}m{2.3cm}|>{\centering\arraybackslash}m{0.6cm}>{\centering\arraybackslash}m{0.6cm}>{\centering\arraybackslash}m{0.6cm}>{\centering\arraybackslash}m{0.8cm}|>{\centering\arraybackslash}m{0.6cm}}
    \toprule
    Size &Model & Visual Encoder & LLM Decoder & DocVQA & InfoVQA & ChartQA & TextVQA & Avg. \\
   \midrule
    \multirow{10}{*}{$<$2B}&DocLLM-1B~\cite{Wang2023ARXIV_DocLLM_A_layout} & -  & Falcon-1B  & 61.4 & - & - & - & -\\
    &Mini-Monkey~\cite{Huang2024ARXIV_Mini_Monkey_Alleviating} & InternViT-300M &InternLLM2-2B  & 87.4 & 60.1 & 76.5 & 75.7 & 74.93\\
    &MM1.5-1B~\cite{zhang2024mm1_5} & CLIP-ViT-H  & Private  & 81.0 & 50.5 & 67.2 & 72.5 &67.80\\
    &MM1.5-3B~\cite{zhang2024mm1_5} & CLIP-ViT-H  & Private  & 87.7 & 58.5 & 74.2 & 76.5 & 74.23\\
    &InternVL2-1B~\cite{InternVL2} & InternViT-300M  & Qwen2-0.5B  & 81.7 & 50.9 & 72.9 & 70.5 & 69.00\\
    &InternVL2-2B~\cite{InternVL2} & InternViT-300M  & InternLM2-1.8B  & 86.9 & 58.9 & 76.2 & 73.3 & 73.83\\
    &InternVL2.5-1B~\cite{chen2024internvl2_5} & InternViT-300M  & Qwen2.5-0.5B  & 84.8 & 56.0 & 75.9 & 72.0 & 72.18\\
    &InternVL2.5-2B~\cite{chen2024internvl2_5} & InternViT-300M  & InternLM2.5-1.8B  & 88.7 & 60.9 & 79.2 & 74.3 & 75.78\\
    &LLaVA-OneVision-0.5B~\cite{li2024llava} & SigLIP  & qwen2-0.5B  & 70.0 & 41.8 & 61.4 & - & - \\
    &\cellcolor{gray!30}TokenVL-2B &\cellcolor{gray!30}TokenFD&\cellcolor{gray!30}InternLM2.5-1.8B&\cellcolor{gray!30}\textbf{89.9}&\cellcolor{gray!30}\textbf{61.0}&\cellcolor{gray!30}\textbf{81.1}&\cellcolor{gray!30}\textbf{76.4}&\cellcolor{gray!30}\textbf{77.10}\\
    \midrule
   \multirow{36}{*}{$<$8B}&UReader~\cite{ye2023ureader} & CLIP-ViT-L/14  & LLaMA-7B  & 65.4 & 42.2 & 59.3 & 57.6 &56.13 \\
    &DocLLM-7B~\cite{Wang2023ARXIV_DocLLM_A_layout} & -  & LLaMA2-7B  & 69.5 & - & - & - &-\\
    &Cream \cite{kim2023visually}& CLIP-ViT-L/14  & Vicuna-7B  & 79.5 & 43.5 & 63.0 & - &- \\
    &Qwen-VL~\cite{bai2023qwenvl} & ViT-bigG  & Qwen-7B  & 65.1 & 35.4 & 65.7 & 63.8 &57.50\\
    &LLaVA-1.5-7B~\cite{liu2023improvedllava} & CLIP-ViT-L  & Vicuna1.5-7B  & - & - & - & 58.2 &- \\
    &SPHINX~\cite{lin2023sphinx} & CLIP-ViT+CLIP-ConvNext+DINOv2-ViT  & LLaMA2-7B  & - & - & - & 61.2 & -\\
    &LLaVA-OneVision~\cite{li2024llavaonevisioneasyvisualtask} & SigLIP  & Qwen2-7B  & 87.5 & 68.8 & 80.0 & - &-\\
    &Monkey~\cite{Li2024CVPR_Monkey_Image_Resolution} & Vit-BigG  & Qwen-7B  & 66.5 & 36.1 & 65.1 & 67.6 & 58.83\\
    &TextMonkey~\cite{Liu2024ARXIV_TextMonkey_An_OCR} & Vit-BigG  & Qwen-7B  & 73.0 & - & 66.9 & 65.6 & -\\
    &IDEFICS2 (\cite{laurençon2024mattersbuildingvisionlanguagemodels}) & SigLIP-SO400M  & Mistral-7B  & 74.0 & - & - & 73.0 & -\\
    &LayoutLLM~\cite{luo2024layoutllm} & LayoutLMv3-large  & Vicuna1.5-7B  & 74.25 & - & - & - & -\\
    &DocKylin~\cite{zhang2024dockylin} & Swin  & Qwen-7B  & 77.3 & 46.6 & 66.8 & - & -\\
    &DocLayLLM~\cite{liao2024doclayllm}  & LayoutLMV3  & LLaMA3-8B  & 77.79 & 42.02 & - & - & -\\
    &mPLUG-DocOwl ~\cite{Hu2024ARXIV_mPLUG_DocOwl_1}  & CLIP-ViT-L/14  & LLaMA-7B  & 62.2 & 38.2 & 57.4 & 52.6 & 52.60\\
    &mPLUG-DocOwl1.5 ~\cite{hu2024mplug_docowl1.5_arxiv}  & CLIP-ViT-L/14   & LLaMA2-7B  & 82.2 & 50.7 & 70.2 & 68.6 & 67.93\\
    &mPLUG-DocOwl2 ~\cite{Hu2024ARXIV_mPLUG_DocOwl2_High}  & CLIP-ViT-L/14   & LLaMA2-7B  & 80.7 & 46.4 & 70.0 & 66.7 & 65.95\\
    &Vary~\cite{wei2024vary}  & CLIP-ViT-L/14 + SAM  & Qwen-7B  & 76.3 & - & 66.1 & - & -\\
    &Eagle~\cite{shi2024eagleexploringdesignspace}  & CLIP + ConvNeX + Pix2Struct + EVA2 + SAM  & LLaMA3-8B  & 86.6 & - & 80.1 & 77.1 & -\\
    &PDF-WuKong~\cite{Xie2024ARXIV_PDF_WuKong_A} & CLIP-ViT-L-14  & InernLM2-7B  & 85.1 & 61.3 & 80.0 & - & -\\
    &TextHawk2~\cite{yu2024texthawk2}  & SigLIP  & Qwen2-7B  & 89.6 & 67.8 & 81.4 & 75.1 &78.48\\
    &MM1.5-7B~\cite{zhang2024mm1_5} & CLIP-ViT-H  & Private  & 88.1 & 59.5 & 78.6 & 76.5 & 75.68\\
    &HRVDA~\cite{liu2024hrvda} & Swin-L  & LLaMA2-7B  & 72.1 & 43.5 & 67.6 & 73.3 & 64.13\\
    &InternVL2-4B~\cite{InternVL2} & InternViT-300M  & Phi-3-mini  & 89.2 & 67.0 & 81.5 & 74.4 & 78.03\\
    &InternVL2-8B~\cite{InternVL2} & InternViT-300M  & InternLM2.5-7B  & 91.6 & 74.8 & 83.3 & 77.4 & 81.78\\
    &InternVL2.5-4B~\cite{chen2024internvl2_5} & InternViT-300M  & Qwen2.5-3B  & 91.6 & 72.1 & 84.0 & 76.8 & 81.13 \\
    &InternVL2.5-8B~\cite{chen2024internvl2_5} & InternViT-300M  & InternLM2.5-7B  & 93.0 & \textbf{77.6} & 84.8 & 79.1 & 83.63 \\
    &InternVL2.5-8B-mpo\cite{wang2024enhancingreasoningabilitymultimodal}$\dagger$  & InternViT-300M  & InternLM2.5-7B & 92.3 & 76.0 & 83.8 & 79.1 & 82.80 \\
    &DeepSeek-VL2-3B~\cite{wu2024deepseek} & SigLIP-SO400M-384  & DeepSeekMoE  &88.9 &66.1 &81.0 &80.7 &79.18 \\
    &DocPeida~\cite{Feng2024ARXIV_DocPedia_Unleashing_the} & Swin  & Vicuna-7B  & 47.1 & 15.2 & 46.9 & 60.2 & 42.35 \\
    &TokenPacker-7B~\cite{li2024tokenpackerefficientvisualprojector} & CLIP-ViT-L/14  & Vicuna-7B  & 60.2 & - & - & - & - \\
    &LLaVA-OneVision-7B~\cite{li2024llava} & SigLIP  & qwen2-7B  & 87.5 & 68.8 & 80.0 & - & - \\
    &DocVLM~\cite{nacson2024docvlmmakevlmefficient} & CLIP-ViT-G/14 + DocFormerV2  & Qwen2-7B  & 92.8 & 66.8 & - & \textbf{82.8} & - \\
    &\cellcolor{gray!30}TokenVL-8B &\cellcolor{gray!30}TokenFD&\cellcolor{gray!30}InternLM2.5-7B&\cellcolor{gray!30}\textbf{94.2}&\cellcolor{gray!30}76.5&\cellcolor{gray!30}\textbf{86.6}&\cellcolor{gray!30}79.9&\cellcolor{gray!30}\textbf{84.30}\\
    \midrule
    \multirow{22}{*}{$>$8B}&LLaVA-13B~\cite{liu2023llava} & CLIP-ViT-L/14  & Vicuna-13B  & 6.9 & - & - & 36.7 & -\\
    &PaLI-X~\cite{chen2023pali} & ViT-22B  & UL2-32B  & 86.8 & 54.8 & 72.3 & 80.8  & 73.68\\
    &LLaVAR~\cite{zhang2023llavar}& CLIP-ViT-L/14  & Vicuna-13B  & 11.6 & - & - & 48.5 & -\\
    &LLaVA-1.5-13B~\cite{liu2023improvedllava} & CLIP-ViT-L  & Vicuna1.5-13B  & - & - & - & 62.5 &- \\
    &CogAgent~\cite{hong2023cogagent} & EVA2-CLIP+CogVLM  +Cross Attention & Vicuna-13B  & 81.6 & 44.5 & 68.4 & 76.1 & 67.65\\
    &Unidoc~\cite{feng2023unidoc} & CLIP-ViT-L/14  & Vicuna-13B  & 90.2 & 36.8 & 70.5 & 73.7 & 67.80 \\
    &MM1.5-30B~\cite{zhang2024mm1_5} & CLIP-ViT-H  & Private  & 91.4 & 67.3 & 83.6 & 79.2 & 80.38\\
   &InternVL1.5-26B~\cite{chen2024far} & InternViT-6B  & InternLM2-20B  & 90.9 & 72.5 & 83.8 & 80.6 & 81.95\\
   &InternVL2-26B~\cite{InternVL2} & InternViT-6B  & InternLM2-20B  & 92.9 & 75.9 & 84.9 & 82.3 & 84.00\\
   &InternVL2-40B~\cite{InternVL2} & InternViT-6B  & Nous-Hermes-2-Yi-34B  & 93.9 & 78.7 & 86.2 & 83.0 & 85.45\\
    &InternVL2.5-26B~\cite{chen2024internvl2_5} & InternViT-6B  & InternLM2.5-20B  & 94.0 & 79.8 & 87.2 & 82.4 & 85.85 \\
    &InternVL2.5-38B~\cite{chen2024internvl2_5} & InternViT-6B  & Qwen2.5-32B  & \textbf{95.3} & 83.6 & 88.2 & 82.7 & 87.45\\
    &InternVL2.5-78B~\cite{chen2024internvl2_5} & InternViT-6B  & Qwen2.5-72B  & 95.1 & \textbf{84.1} & \textbf{88.3} & 83.4 & \textbf{87.73} \\
    &TinyChart~\cite{zhang2024tinychart} & SigLIP  & Phi-2  & - & - & 83.6 & - & - \\
    &TokenPacker-13B~\cite{li2024tokenpackerefficientvisualprojector} & CLIP-ViT-G/14  & Vicuna-13B  & 70.0 & - & - & - & - \\
    &DeepSeek-VL2-16B~\cite{wu2024deepseek} & SigLIP-SO400M-384  & DeepSeekMoE  &92.3 &75.8 &84.5 &83.4 &84.00 \\
    &DeepSeek-VL2-27B~\cite{wu2024deepseek} & SigLIP-SO400M-384  & DeepSeekMoE  &93.3 &78.1 &86.0 &\textbf{84.2} &85.40 \\
    \bottomrule
    \end{tabular}
    \caption{Comparison results on four widely evaluated datasets.}
    \label{tab:popular}
\end{table}

\end{document}